\documentclass[]{style}

\usepackage[toc,page,header]{appendix}
\usepackage{enumitem}
\usepackage{xcolor}
\usepackage[dvipsnames]{xcolor}
\usepackage{listings}
\usepackage{caption}

\usepackage{natbib}

\usepackage{CJKutf8}

\usepackage{xargs}  

\usepackage{todonotes}  

\usepackage{multirow}

\usepackage{cleveref}

\usepackage{amsmath}
\usepackage{dsfont}

\usepackage{svg}

\usepackage{mathrsfs}
\usepackage{adjustbox}
\usepackage{multirow}
\usepackage{multicol}
\usepackage{tcolorbox}
\usepackage{changepage}
\usepackage{enumitem}
\usepackage{graphicx}
\usepackage{amssymb}
\usepackage{xcolor}
\usepackage{float}
\usepackage{multirow}
\usepackage{threeparttable}
\usepackage{graphicx}
\usepackage{subcaption}
\usepackage{algorithm}
\usepackage{algpseudocode}
\usepackage{wrapfig}
\usepackage[table]{xcolor}  
\usepackage{colortbl}       
\usepackage{tabularx} 
\usepackage{makecell} 
\usepackage[dvipsnames]{xcolor}

\newcolumntype{g}{>{\columncolor{gray!10}}c} 

\definecolor{catgray}{gray}{0.9}
\definecolor{skyblue}{rgb}{0.53,0.81,0.92} 

\colorlet{skyblue!30}{skyblue!30!white} 

\definecolor{customblue}{RGB}{70,130,180}  

\newtcolorbox{evolbox}[2][]{%
  enhanced,
  colframe=customblue,
  colback=white,
  coltitle=white,
  rounded corners,
  boxrule=1pt,
  titlerule=0pt,
  toptitle=1mm,
  bottomtitle=1mm,
  fonttitle=\bfseries,
  width=#2\textwidth, 
  #1
}

\usepackage{minitoc}
\usepackage{url}

\PassOptionsToPackage{table,xcdraw}{xcolor}
\usepackage{titletoc}
\usepackage{placeins}
\usepackage{pifont}

\definecolor{RowBlue}{HTML}{E9F2FB}
\definecolor{RowRed}{HTML}{F9EAEA}
\definecolor{Top1}{HTML}{50DB4B} 
\definecolor{Top2}{HTML}{A5FFA2} 
\definecolor{Top3}{HTML}{D9FFD9} 
\definecolor{Sub1}{HTML}{EAB8B8}
\definecolor{Sub2}{HTML}{E4E4E4}

\renewcommand{\emph}[1]{\textit{#1}}

\definecolor{codepink}{RGB}{220,20,120}
\definecolor{codegreen}{RGB}{0,150,0}
\definecolor{codegray}{RGB}{140,140,140}
\definecolor{codeorange}{RGB}{230,120,60}

\lstdefinestyle{pytorchstyle}{
    language=Python,
    basicstyle=\ttfamily\small,
    keywordstyle=\color{codepink}\bfseries,
    commentstyle=\color{codegray}\itshape,
    stringstyle=\color{codeorange},
    numberstyle=\tiny\color{codegray},
    numbers=none,
    showstringspaces=false,
    breaklines=true,
    frame=none,
    columns=fullflexible,
    keepspaces=true,
    xleftmargin=1.5em
}

\title{OSP-Next: Efficient High-Quality Video Generation with Sparse Sequence Parallelism, HiF8 Quantization, and Reinforcement Learning}

\author{
  Yunyang Ge\texorpdfstring{$^{1,*}$}{} \hspace{0.1cm}
  Xianyi He\texorpdfstring{$^{1,*}$}{} \hspace{0.1cm}
  Zezhong Zhang\texorpdfstring{$^{2}$}{} \hspace{0.1cm}
  Bin Lin\texorpdfstring{$^{1}$}{} \\
  Bin Zhu\texorpdfstring{$^{1,3}$}{} \hspace{0.1cm}
  Xinhua Cheng\texorpdfstring{$^{1}$}{} \hspace{0.1cm}
  Li Yuan\texorpdfstring{$^{1,\dag}$}{} \hspace{0.1cm}
}

\makeatletter
\g@addto@macro\authorlist{\\[2mm]
  {\authorfont\sffamily
  Open-Sora Plan Team
  }\\[1.2mm]
  {\small
  \texorpdfstring{$^{*}$}{}Equal contribution, \quad
  \texorpdfstring{$^{\dag}$}{}Corresponding author
  }\\[1.2mm]
  {\small\ttfamily
  \{yunyang, HeXianyi, linbin.ece, binzhu, chengxinhua\}@stu.pku.edu.cn
  }\\[-0.2mm]
  {\small\ttfamily
  zezhong002@e.ntu.edu.sg
  }
  {\small\ttfamily
  yuanli-ece@pku.edu.cn
  }
}
\makeatother

\affiliation[1]{Peking University}
\affiliation[2]{Nanyang Technological University, Singapore}
\affiliation[3]{Rabbitpre AI}

\newcommand{\answerTODO}[1][]{\textcolor{red}{\bfseries [TODO]}}
\newcommand{\justificationTODO}[1][]{\textcolor{red}{\bfseries [TODO]}}

\abstract{
Diffusion Transformers achieve strong video generation quality, but the quadratic cost of full attention limits efficiency. We introduce \textbf{OSP-Next}, an efficient text-to-video generation model that integrates sparse attention, parallelism, quantization, and reinforcement learning. OSP-Next uses a hybrid full-sparse attention architecture, where the sparse component is implemented with \textbf{Skiparse-2D Attention}. This fixed-pattern mechanism applies token-wise and group-wise sparse attention along spatial dimensions, leveraging locality while maintaining \textbf{native compatibility with FlashAttention kernels}. Based on the local equivalence of rearrangement in Skiparse-2D Attention, we further propose \textbf{Sparse Sequence Parallelism (SSP)}, which partitions subsequences across ranks and switches sparse patterns through a single All-to-All communication. Compared with Ulysses Sequence Parallelism (SP), SSP provides a \textbf{native parallel strategy for sparse attention} and reduces communication volume by \textbf{75\%}. OSP-Next also incorporates HiF8 quantization to enable stable joint training with 8-bit quantization and sparse fine-tuning, and applies Mix-GRPO post-training to improve the performance of the sparse model. Experiments show that OSP-Next achieves a \textbf{VBench total score of 83.73\%, surpassing the Wan2.1 baseline}. Under the 5-second 720P and 5-second 768P settings, OSP-Next achieves up to \textbf{1.64$\times$ single-GPU speedup} and over \textbf{1.52$\times$ eight-GPU speedup} on NVIDIA H200 GPUs. In addition, with only a 0.4\% drop in VBench total score, OSP-Next-HiF8 achieves \textbf{1.69$\times$ and 2.27$\times$ speedups} under the two settings on a single Ascend 950PR, demonstrating the efficiency and performance of OSP-Next across hardware platforms.
}

\checkdata[
\raisebox{-0.2em}{\includegraphics[width=0.025\linewidth]{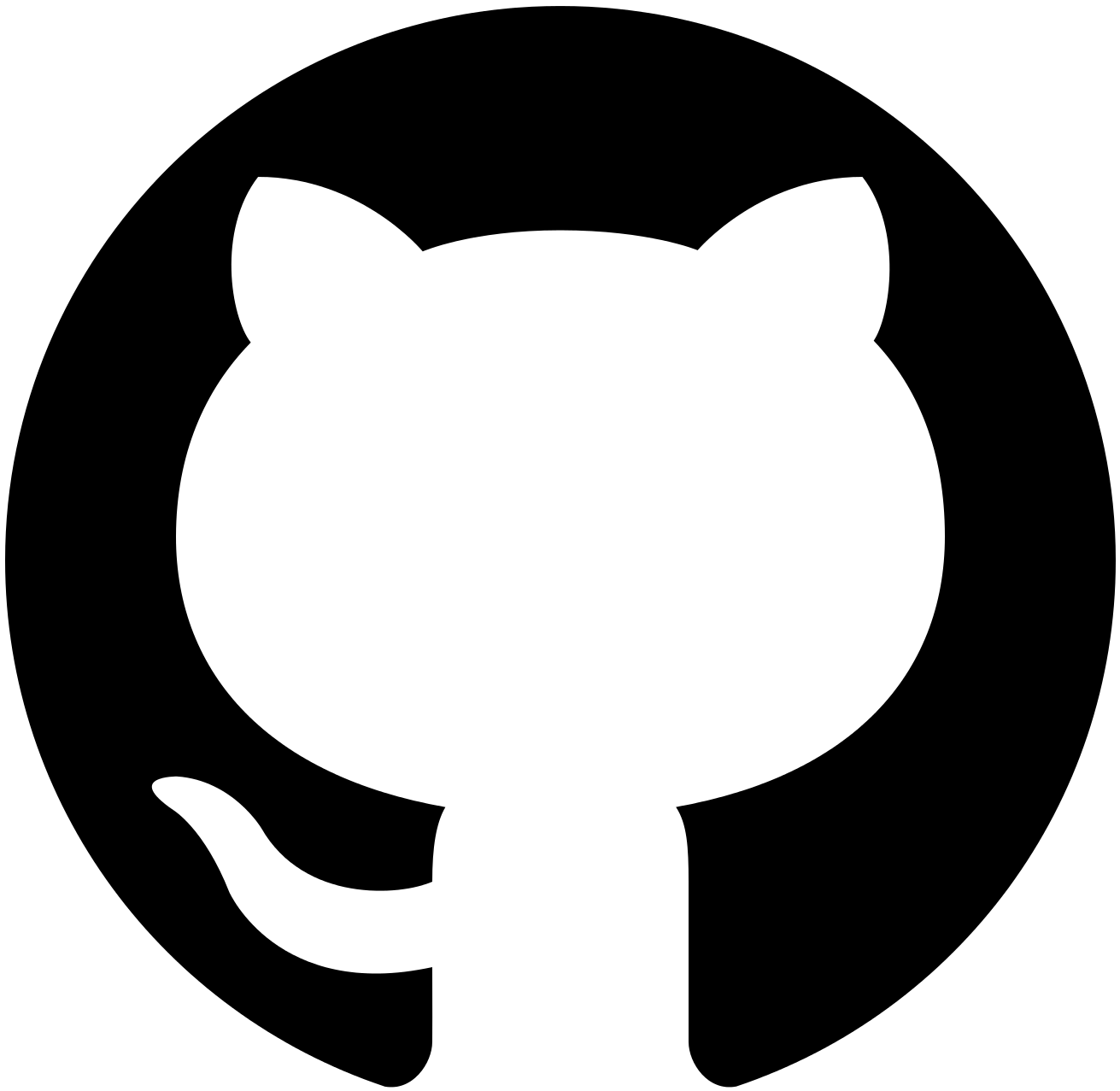}}~~GitHub Repo]{\href{https://github.com/PKU-YuanGroup/OSP-Next}{\texttt{https://github.com/PKU-YuanGroup/OSP-Next}}
\\[-1.7ex]}

\checkdata[
\raisebox{-0.3em}{\includegraphics[width=0.026\linewidth]{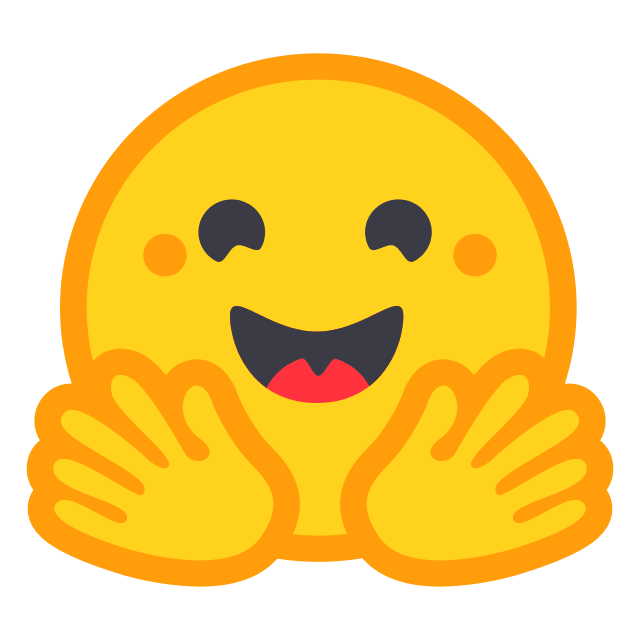}}~~HuggingFace Model]{\href{https://huggingface.co/yunyangge/OSP-Next}{\texttt{https://huggingface.co/yunyangge/OSP-Next}}
\\[-1.7ex]}

\begin{document}
\maketitle

\section{Introduction}
\label{sec:intro}
In recent years, Diffusion Transformers (DiTs)~\citep{dit,lightingdit} have increasingly replaced UNets~\citep{unet,stable_diffusion} as denoisers for image and video generation~\citep{stable_diffusion_3,latte,cogvideox,open_sora,open_sora_2,open_sora_plan,hunyuanvideo,hunyuanvideo1.5,wan2.1,flashi2v}, demonstrating strong generative capacity. However, the computational inefficiency of DiTs and the long token sequences required for video modeling limits speed and achievable resolution. For example, models such as HunyuanVideo~\citep{hunyuanvideo} and Wan2.1~\citep{wan2.1} require approximately 30 minutes to one hour to generate a 5-second 720p video on a single NVIDIA A100 GPU.

To mitigate the challenges introduced by quadratic computational complexity, prior works have explored various techniques, including sparse attention, linear attention, and quantization. Training-free sparse attention methods~\citep{sparsevideogen,sparsevideogen2,AdaSpa,spargeattention} typically select tokens dynamically according to token similarity scores. This strategy relies on a pretrained full attention model and keeps the model weights fixed, preventing the model from adapting to the distribution shift caused by changes in the attention pattern. Training-based sparse attention methods~\citep{sta,sparse_vdit} also commonly select tokens according to predefined similarity rules. However, these methods require dynamically constructed attention masks and are typically implemented with FlexAttention~\citep{flexattention} or custom kernels specifically optimized for sparse attention, making them difficult to natively integrate with efficient FlashAttention~\citep{flashattention,flashattention2,flashattention3} kernels. Moreover, irregular attention masks are difficult to parallelize and can lead to load imbalance across different ranks. Linear attention~\citep{sanavideo,sla,sla2} reduces computational complexity to linear in the sequence length by design, but this simplification also limits the expressive capacity of the model. Quantization methods, including FP8 and INT8, provide a substantially narrower representational range than BF16 and FP16. Consequently, models trained with 8-bit precision~\citep{sageattention,turbodiffusion} often exhibit limited performance.

We introduce \textbf{OSP-Next}, a text-to-video model that integrates sparse attention, parallelism, quantization, and reinforcement learning. Through a combination of optimization strategies, \textbf{OSP-Next} mitigates the performance and efficiency issues observed in prior methods while satisfying the requirements of parallel computation and native compatibility with FlashAttention kernels. Our main contributions are as follows.

\begin{enumerate}[leftmargin=*, labelsep=0.5em]
\item We propose \textbf{Skiparse-2D Attention}. In contrast to sparse patterns based on dynamic token selection, Open-Sora Plan v1.3 introduces Skiparse-1D Attention, a fixed-rule attention pattern that alternates between token-wise skip sparse attention and group-wise skip sparse attention to approximate 3D full attention. In OSP-Next, Skiparse Attention is applied separately along the height and width dimensions to form Skiparse-2D Attention, which better aligns with the spatial locality of image and video modalities. Experimental results indicate that Skiparse-2D Attention captures an interaction pattern closer to that of 3D full attention than Skiparse-1D Attention, thereby achieving better performance. Moreover, by avoiding dynamic construction of a 2D attention mask over the sequence dimension, Skiparse-2D Attention remains natively compatible with efficient FlashAttention kernels without custom kernels.
\item We introduce \textbf{Sparse Sequence Parallelism (SSP)}. Skiparse Attention first performs Skiparse Rearrange, which partitions the original sequence into multiple subsequences concatenated along the batch dimension, and then computes attention in parallel. Skiparse Rearrange exhibits \textbf{local equivalence}: applying Skiparse Rearrange to the original sequence is equivalent to applying it independently to each minimal repeatable unit. Therefore, different subsequences can be assigned to different ranks, with attention computation performed independently on each rank. When the sparse pattern of Skiparse Attention switches between token-wise sparse attention and group-wise sparse attention, the local equivalence property of Skiparse Rearrange allows data exchange to be completed by one local rearrangement on each rank followed by one All-to-All communication step within the communication group. In addition, since the subsequences produced by Skiparse Rearrange have equal lengths, load balance is maintained across ranks. Compared with Ulysses Sequence Parallelism (SP), SSP reduces the communication volume by \textbf{75\%} and decreases the number of communication steps per block from four to one.
\item We introduce \textbf{joint training of sparse-model fine-tuning and HiF8 quantization}. HiF8 is an 8-bit precision format that dynamically adjusts the numbers of exponent and mantissa bits, enabling both high precision and a large dynamic range. Benefiting from the utilization of locality in image and video modalities by Skiparse-2D Attention and the training stability of the hybrid architecture in OSP-Next, which combines full attention Blocks and Skiparse Attention Blocks, we can perform sparse-model fine-tuning and 8-bit fine-tuning simultaneously. The loss curve of OSP-Next trained with HiF8 almost overlaps with that of OSP-Next trained with BF16, and the gap in VBench~\citep{vbench} score remains within 0.5\%.
\item We introduce \textbf{reinforcement learning for sparse models}. Although the attention pattern of OSP-Next closely approximates that of the pretrained full-attention model, fine-tuning from the pretrained model to the sparse model still degrades generation quality. To mitigate this quality degradation, we further optimize OSP-Next with Mix-GRPO~\citep{li2025mixgrpo} during the post-training stage. Experimental results show that reinforcement learning substantially improves the generation quality of sparse models.
\end{enumerate}

OSP-Next achieves a \textbf{VBench total score of 83.73\%}, outperforming the Wan2.1 baseline based on full attention. Under the 5-second 720P with padding and 5-second 768P without padding settings, OSP-Next achieves 1.53$\times$ and 1.64$\times$ speedups on a single NVIDIA H200, and 1.42$\times$ and 1.52$\times$ speedups on eight NVIDIA H200 GPUs, respectively. In addition, OSP-Next-HiF8 achieves $1.69\times$ and $2.27\times$ speedups on a single Ascend 950PR, with only a 0.4\% VBench total score drop relative to the baseline. These results show that OSP-Next provides high-quality generation and acceleration across different hardware platforms, offering a new pipeline for trainable and parallelizable sparse video generation models.
\section{Related Work}
\label{sec:related_work}
\subsection{Sparse Video Generation Model}
Sparse video generation models have recently attracted increasing attention as a practical solution to the high computational cost of video Diffusion Transformers. Open-Sora Plan~\citep{open_sora_plan} introduces Skiparse Attention for video generation, which sparsifies 3D attention by reorganizing tokens into sparse subsequences while preserving spatio-temporal modeling ability. Sparse VideoGen~\citep{sparsevideogen} accelerates pretrained video DiTs in a training-free manner by identifying spatial and temporal attention heads and applying head-wise sparse computation during inference. Sparse VideoGen2~\citep{sparsevideogen2} further improves training-free sparse inference by using semantic-aware token permutation, which clusters related tokens to obtain more efficient sparse attention layouts. Sparse-vDiT~\citep{sparse_vdit} analyzes the attention maps of video DiTs and exploits recurring sparse patterns, such as diagonal and stripe structures, with hardware-aware sparse kernel selection. VSA~\citep{vsa} adopts a trainable coarse-to-fine sparsification strategy, where video tokens are first grouped into tiles and only high-importance tiles are selected for fine-grained token-level attention. Currently, trainable sparse methods for VideoDiT remain limited and cannot provide speed gains during pretraining. Moreover, existing trainable sparse methods are mostly based on complex dynamic token selection strategies. Such methods are difficult to combine with parallelization strategies and usually require complex masks implemented with FlexAttention, making them incompatible with flash attention.

\subsection{Parallel Strategy}
Scaling video generation models requires efficient parallel training and inference strategies. 
Data parallelism, such as DDP, replicates the full model on each device and splits the input batch, but does not reduce the memory cost of model parameters or activations on each GPU. 
Tensor parallelism~\citep{megatron} partitions matrix operations or attention heads across devices, reducing per-device computation but introducing communication within Transformer layers. 
FSDP~\citep{fsdp} further shards model parameters, gradients, and optimizer states across devices, making it effective for training large models under limited GPU memory. 
However, these strategies mainly parallelize the batch, channel, or parameter dimensions, while video generation is often bottlenecked by the extremely long spatio-temporal sequence length.

Sequence parallelism is therefore particularly important for video Diffusion Transformers. 
Ulysses-style sequence parallelism~\citep{ulysses_sp} partitions the sequence dimension and uses All-to-All communication to gather the required tokens for attention computation. 
Ring Attention~\citep{ringattention} instead computes attention in a blockwise manner by circulating key-value blocks among devices, enabling long-sequence attention while overlapping communication with computation. 
Recent unified sequence-parallel frameworks~\citep{usp} further combine these two paradigms to improve scalability under different sequence lengths and hardware settings.

Despite these advances, the interaction between sequence parallelism and sparse video generation remains underexplored. 
Most sparse video generation methods focus on designing sparse attention patterns or accelerating pretrained models, but do not explicitly consider how the sparse token layout should be distributed across devices. 
Directly combining sparse attention with sequence parallelism may lead to workload imbalance and irregular communication, since different devices can receive different numbers of active tokens or attention blocks. 
Therefore, designing sparse video generation models that are naturally compatible with sequence-parallel execution remains an important and open direction.

\subsection{Fine-Grained Quantization}
A central challenge in FP8~\citep{fp8} mixed-precision training is controlling quantization error caused by the limited dynamic range of standard FP8 formats, such as E4M3 and E5M2. 
To mitigate this issue, existing methods commonly adopt fine-grained quantization, where independent scaling factors are assigned to smaller subsets of tensors to better match local numerical distributions.

For activations, per-token quantization assigns an individual scale to each token or tile. 
This is effective for DiT-based models and LLMs, where activation magnitudes vary significantly across tokens and are often affected by outliers. 
Compared with per-tensor scaling, per-token scaling prevents a few outlier tokens from dominating the global scale and thus reduces quantization error. 
However, it requires computing token-wise maximum values before quantization, introducing reduction overhead and creating a serial dependency before MatMul operations.

For weights, per-channel and per-block quantization are widely used to capture distribution differences across output channels or local matrix regions. 
Per-channel quantization assigns one scale to each output channel, while per-block quantization further divides the weight matrix into fixed-size tiles, such as $128\times128$, to balance accuracy and metadata cost. 
More fine-grained schemes, such as MXFP8~\citep{mxfp8} microscaling, assign a shared microscale to every 32 consecutive elements, offering stronger robustness to local outliers but requiring denser scale metadata and more complex hardware support.

Overall, fine-grained quantization improves numerical accuracy by using more adaptive scaling, but this comes with non-negligible system overhead. 
Frequent scale computation, storage, loading, and broadcasting increase memory bandwidth pressure and complicate hardware implementation, partially offsetting the efficiency benefits of 8-bit computation. 
In contrast, HiF8~\citep{hif8} provides a wider dynamic range and can therefore rely on coarse per-tensor quantization while maintaining training accuracy close to the BF16 baseline. 
This avoids much of the metadata and reduction overhead required by fine-grained FP8 quantization, leading to a simpler and more efficient training pipeline.

\subsection{Reinforcement Learning for Video Generation}
Reinforcement Learning (RL) post-training has recently been explored to improve the preference alignment of diffusion-based video generation models by directly optimizing reward signals related to visual quality, motion consistency, and text-video alignment. 
FlowGRPO~\citep{liu2026flow} formulates the denoising process as a Markov decision process and applies GRPO-style optimization over the full trajectory, but full-step optimization is costly for video generation due to long spatio-temporal sequences and expensive video-level reward evaluation. 
Mix-GRPO~\citep{li2025mixgrpo} improves efficiency by combining stochastic SDE sampling with deterministic ODE sampling, restricting policy-gradient updates to selected denoising timesteps. 
Following this trajectory-level RL paradigm, DanceGRPO~\citep{xue2025dancegrpo} adapts GRPO to dance video generation with rewards that emphasize pose dynamics, motion naturalness, and rhythm consistency, while BranchGRPO~\citep{li2025branchgrpo} introduces branched denoising rollouts, allowing multiple candidates to share early denoising computation before diverging for reward-based optimization.

Different from these GRPO-style methods that optimize sampled denoising trajectories with policy-gradient updates, DiffusionNFT~\citep{zheng2025diffusionnft} follows a different reward fine-tuning paradigm for diffusion models. 
It avoids relying on approximate diffusion transition probabilities for importance sampling or policy-ratio estimation, which are commonly used when casting the denoising process as an MDP. 
By reducing the dependence on such transition approximation, DiffusionNFT mitigates potential bias in diffusion RL post-training and provides a more direct optimization route. 
For video generation, this is especially relevant because trajectory-level rollout, policy-ratio computation, and reward evaluation become particularly expensive for long spatio-temporal sequences.
Despite these advances, existing RL post-training methods mainly target standard diffusion or video generation models, while their compatibility with sparse attention, sequence parallelism, and low-precision training remains less studied.
\section{Method}\label{sec:dataset}
\subsection{Skiparse-2D Attention}
\begin{figure}[t]
    \centering
    \includegraphics[width=0.8\linewidth]{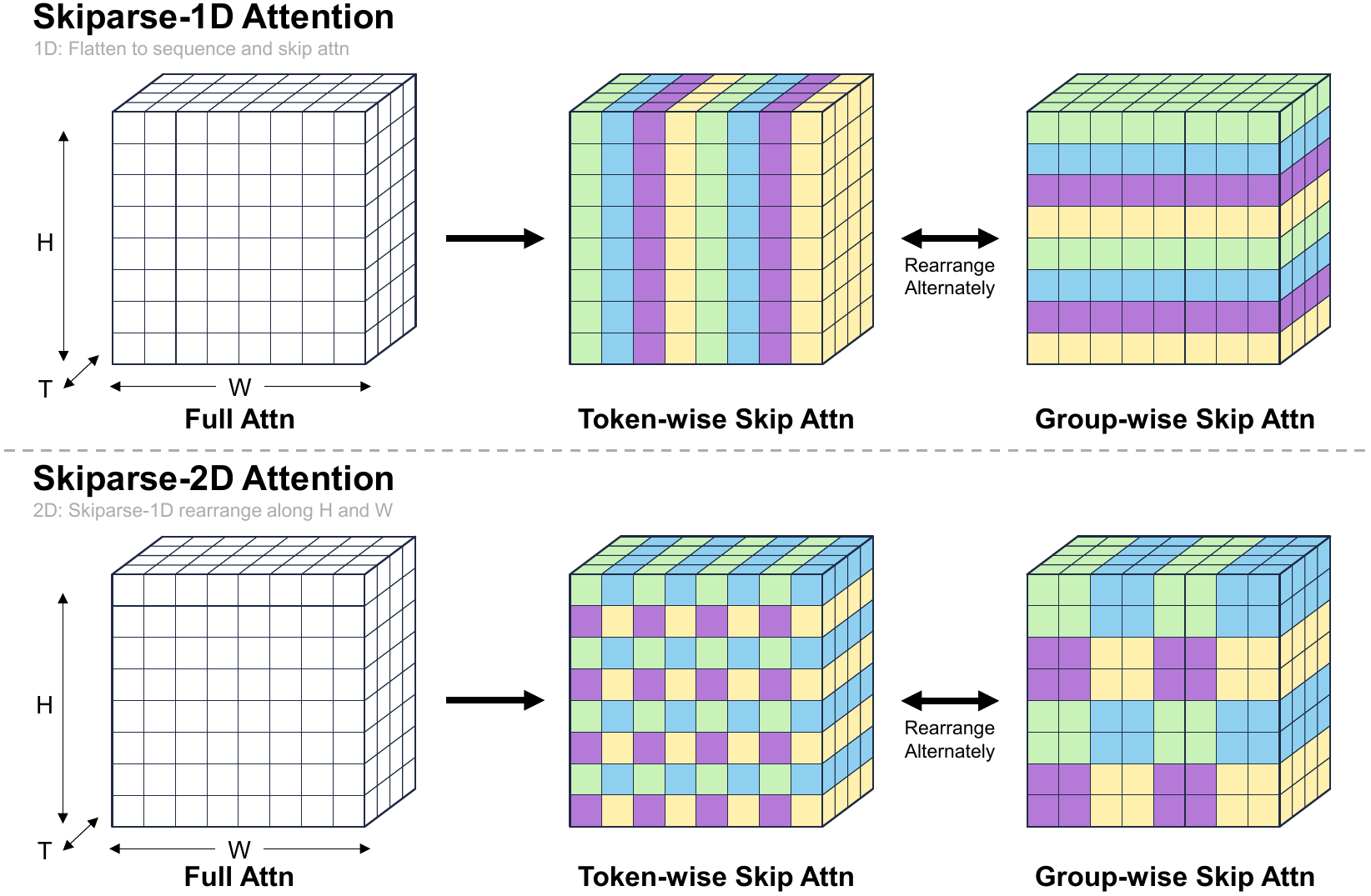}
    \caption{\textbf{Skiparse-2D Attention.} Skiparse-2D Attention alternates between Token-wise Sparse Attention and Group-wise Sparse Attention, enabling any two tokens to interact within at most two attention operations. Unlike Skiparse-1D Attention, which treats a video as a flattened one-dimensional sequence, Skiparse-2D Attention applies the Skiparse-1D pattern orthogonally along the height and width dimensions, thereby better leveraging the locality of image and video modalities.}
    \label{fig:method_skiparse_2d_attn}
\end{figure}
Open-Sora Plan v1.3~\citep{open_sora_plan} first introduces Skiparse Attention, a trainable sparse attention mechanism for DiTs. In this work, video latents are flattened into a one-dimensional sequence, and token-wise skip sparse attention and group-wise skip sparse attention are applied alternately to construct a sparse attention mechanism whose computational cost lies between spatial-temporal attention and 3D full attention. Open-Sora Plan v1.3 adopts a hybrid DiT architecture, applying full attention in the first and last several layers while using Skiparse Attention in the middle layers, which provides substantial acceleration while maintaining sufficient generation quality. In the subsequent Open-Sora Plan v1.5, the model is extended to a hybrid architecture with a U-shaped varying sparsity pattern. This design achieves a VBench score above 83\% and becomes the first model trained from scratch to approach the performance of open-source 3D full attention models like HunyuanVideo.

However, simply flattening a video into a one-dimensional sequence introduces several issues. First, it is inconsistent with the 2D locality of image and video modalities. Under the above Skiparse-1D Attention, interactions among local tokens are restricted: some spatially adjacent tokens cannot interact within a single attention operation and instead require intermediate tokens. Moreover, the interaction pattern among global tokens fails to match the 2D structure of image and video modalities. These limitations weaken the modeling capability of Skiparse-1D Attention. This limitation becomes more pronounced at higher sparsity levels. Furthermore, in the any resolution training setting, Skiparse-1D Attention can produce inconsistent token interaction patterns. It constructs subsequences from the flattened one-dimensional sequence and pads only the end when necessary, while ignoring the spatial positions of pixels. Consequently, tokens at the same spatial position across different videos may be assigned to different subsequences and thus follow different interaction patterns, further making the model more difficult to optimize.

Therefore, we extend Skiparse-2D Attention to better support image and video modalities. By viewing Skiparse-2D Attention as applying Skiparse-1D Attention separately along the width and height dimensions, we obtain the interactive pattern of Skiparse-2D Attention, as shown in Fig.~\ref{fig:method_skiparse_2d_attn}. The token-wise interaction pattern resembles pixel unshuffle, but the resulting subfigures are concatenated along the batch dimension as different subsequences rather than along the channel dimension. The group-wise interaction resembles patch unshuffle, where the resulting subfigures are likewise treated as different subsequences and concatenated along the batch dimension. During attention computation, interactions occur only within each subfigure. Since the complexity of attention is $O(n^2)$, both token-wise sparse attention and group-wise sparse attention reduce the sequence dimension to $\frac{1}{k}$ of the original size and increase the batch dimension $k$ times. As a result, the total computation of the attention operation is reduced to $\frac{1}{k}$ of the original cost, where $k$ denotes the sparse ratio and corresponds to the skip interval in both operations. By alternating token-wise sparse attention and group-wise sparse attention in the model, any two tokens can interact through at most two attention operations.
\subsection{Any Resolution Strategy}
\begin{figure}[t]
    \centering
    \includegraphics[width=\linewidth]{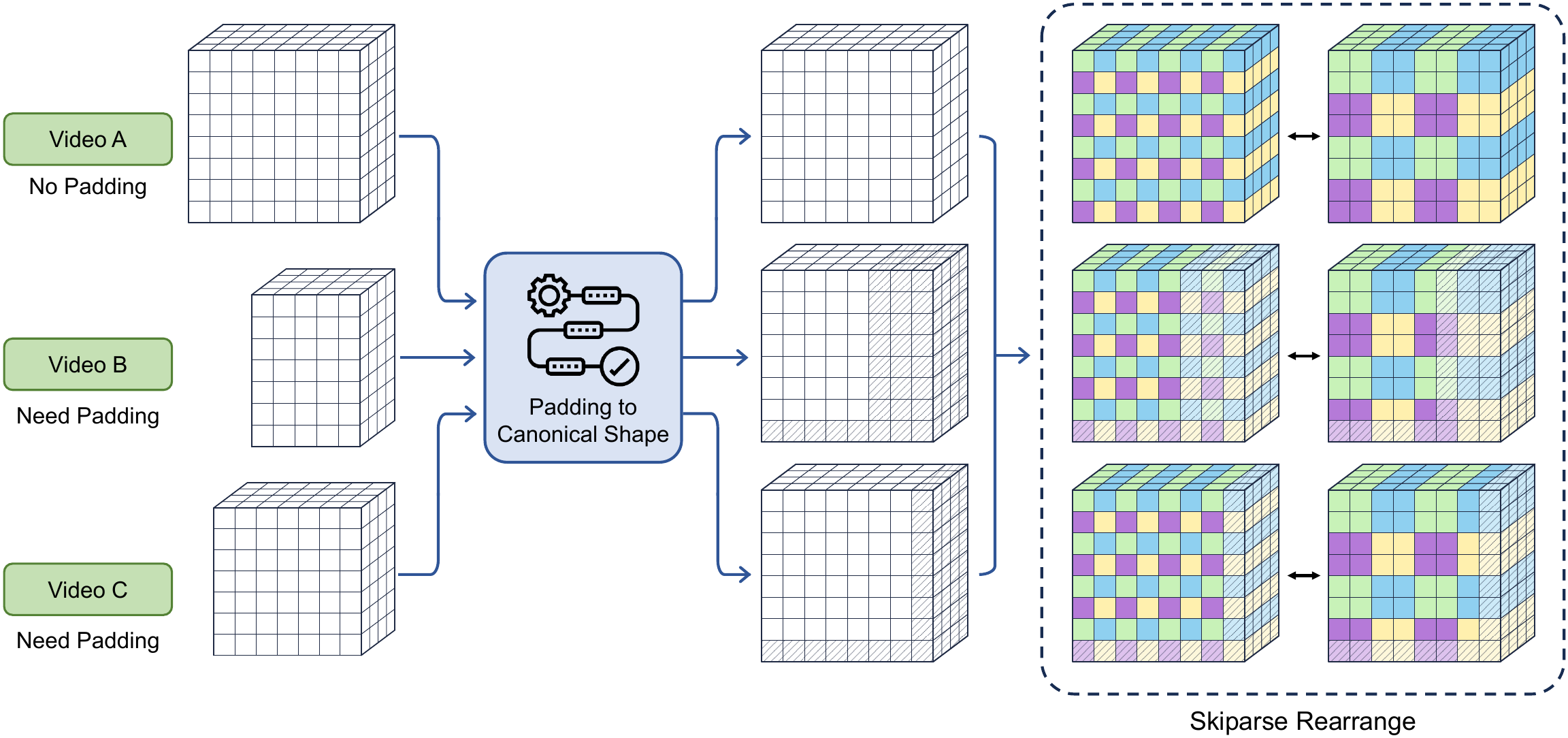}
    \caption{\textbf{Any Resolution Strategy.} The minimum repeating unit of Skiparse-2D Attention is a $k^2 * k^2$ subfigure, where $k$ denotes the sparse ratio. When the height or width of a video shape is not a multiple of $k^2$, we pad it to the nearest multiple of $k^2$ and generate the corresponding attention mask. This any resolution strategy ensures that tokens at the same spatial position across videos of different shapes are assigned to the same subfigure.}
    \label{fig:method_any_resolution}
\end{figure}
The Skiparse Rearrange satisfies local equivalence: applying token-wise or group-wise rearrange to an $H \times W$ figure is equivalent to independently applying the rearrange to $\frac{H}{k^2} \times \frac{W}{k^2}$ subfigures, each of size $k^2 \times k^2$. In other words, the smallest repeatable unit of the Skiparse Rearrange is a $k^2 \times k^2$ subfigure.

When $H$ or $W$ is not divisible by $k^2$, the naive Skiparse Rearrange becomes invalid because some subfigures cannot form the smallest repeatable unit. To support Skiparse Rearrange at arbitrary resolutions, we pad the original figure. The padding operation should preserve the local equivalence of Skiparse Rearrange. Therefore, we apply padding to the end of both the $H$ and $W$ dimensions and construct the corresponding attention mask. As shown in the fig.~\ref{fig:method_any_resolution}, this padding strategy preserves the local equivalence of Skiparse Rearrange and keeps the interaction pattern among non-padding tokens unchanged, thereby enabling training at arbitrary resolutions.

According to the any resolution strategy, the attention mask in Skiparse Attention is set to None when both $H$ and $W$ are exact multiples of $k^2$. Otherwise, it is only a 1D mask of the flattened sequence dimension, without requiring a complex 2D mask. This means that Skiparse Attention natively supports FlashAttention kernels and avoids the compilation overhead of FlexAttention required by other sparse attention methods.
\definecolor{codebg}{RGB}{255,250,232} 

\lstdefinestyle{algocode}{
    language=Python,
    basicstyle=\ttfamily\scriptsize,
    breaklines=true,
    breakatwhitespace=false,
    columns=flexible,
    keepspaces=true,
    showstringspaces=false,
    tabsize=4,
    backgroundcolor=\color{codebg},
    keywordstyle=\bfseries\color{NavyBlue},
    commentstyle=\itshape\color{black!50!white},
    stringstyle=\color{PineGreen!80!black},
    numbers=none,
    frame=single,
    framerule=0pt,
    rulecolor=\color{codebg},
    framesep=0.75em,
    xleftmargin=0pt,
    xrightmargin=0pt,
    framexleftmargin=0pt,
    framexrightmargin=0pt,
    linewidth=\linewidth,
    aboveskip=0.15em,
    belowskip=0.15em
}

\begin{figure}[!t]
\centering
\begin{minipage}[t]{0.45\linewidth}
\vspace{0pt}
\centering
\includegraphics[width=\linewidth]{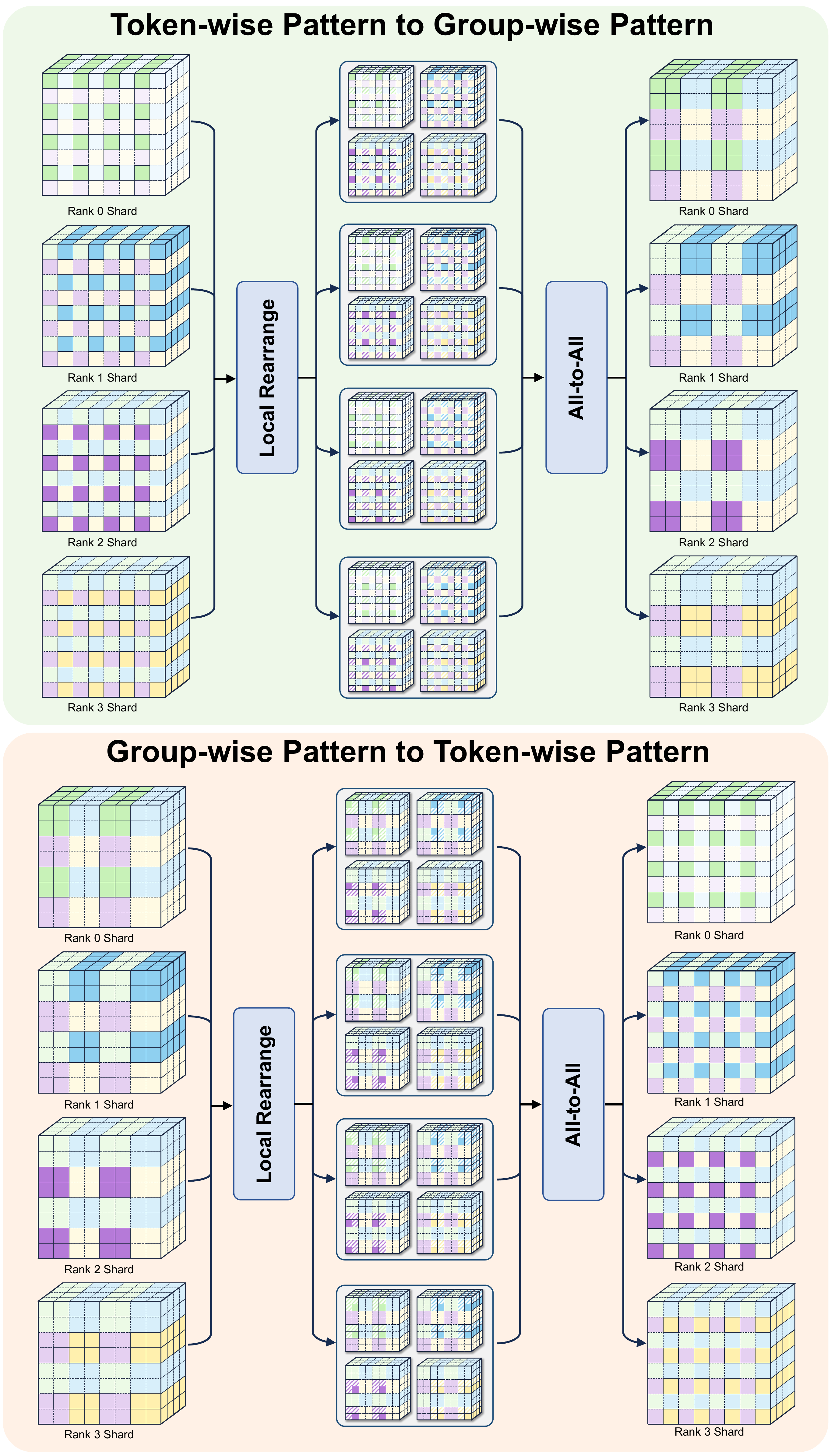}
\captionof{figure}{\textbf{Sparse Sequence Parallel (SSP).} Benefiting from the local equivalence of Skiparse-2D Attention, we can locally rearrange the subfigures sharded to each rank to identify the elements that remain on the same local rank after pattern switching. The target layout can then be obtained through All-to-All communication within the sparse parallel group, without requiring global All-Gather. Compared with Ulysses sequence parallelism (SP), SSP reduces the communication volume by 75\% and requires only one communication operation per block.}
\label{fig:method_ssp}
\end{minipage}
\hfill
\begin{minipage}[t]{0.53\linewidth}
\vspace{0pt}
\setlength{\parindent}{0pt}

\noindent\hrule height 0.45pt
\vspace{0.14em}

\refstepcounter{algorithm}
\label{alg:ssp}
{\small
\noindent\textbf{Algorithm~\thealgorithm}\quad
Pseudocode of Skiparse Rerrange
\par}

\vspace{0.14em}
\noindent\hrule height 0.35pt
\vspace{0.32em}

\begin{lstlisting}[style=algocode]
def orig_to_tsa(x, T, H, W, P):
    return rearrange(x, "b (t h p w q) c -> (p q b) (t h w) c",
                     p=P, q=P, h=H // P, w=W // P)

def tsa_to_orig(x, T, H, W, P):
    return rearrange(x, "(p q b) (t h w) c -> b (t h p w q) c",
                     p=P, q=P, h=H // P, w=W // P)

def orig_to_gsa(x, T, H, W, P):
    return rearrange(
        x, "b (txh p1 p2 w q1 q2) c -> (p1 q1 b) (txh p2 w q2) c",
        p1=P, q1=P, p2=P, q2=P, w=W // (P ** 2)
    )

def gsa_to_orig(x, T, H, W, P):
    return rearrange(
        x, "(p1 q1 b) (txh p2 w q2) c -> b (txh p1 p2 w q1 q2) c",
        p1=P, q1=P, p2=P, q2=P, w=W // (P ** 2)
    )

def tsa_to_gsa(x, T, H, W, P):
    return rearrange(
        x, "(p2 q2 b) (txh_p1 p1 w_q1 q1) c -> (p1 q1 b) (txh_p1 p2 w_q1 q2) c",
        p1=P, q1=P, p2=P, q2=P, w_q1=W // (P ** 2)
    )

def gsa_to_tsa(x, T, H, W, P):
    return rearrange(
        x, "(p1 q1 b) (txh_p1 p2 w_q1 q2) c -> (p2 q2 b) (txh_p1 p1 w_q1 q1) c",
        p1=P, q1=P, p2=P, q2=P, w_q1=W // (P ** 2)
    )

def sparse_sequence_parallel(x, grid_sizes, sparse_ratio, group):
    group_size = get_size(group)
    T, H, W = grid_sizes
    P = sparse_ratio
    P2 = P * P
    G = P2 // group_size
    B_local = x.shape[0]
    b = B_local // G

    # 1. Local rearrangement
    x = orig_to_tsa(x, T, H // P, W // P, P)
    base, C = x.shape[1], x.shape[2]
    # 2. Distributed all-to-all transpose
    x = contiguous(x).view(group_size, G, G * b, base, C)
    recv = empty_like(x)
    all_to_all_single(recv, x, group=group)
    # 3. Local permutation
    recv = recv.view(group_size, G, G, b, base, C)
    recv = recv.permute(0, 2, 1, 3, 4, 5)
    x = contiguous(recv).view(P2 * G * b, base, C)
    # 4. Reverse local rearrangement
    x = tsa_to_orig(x, T, H // P, W // P, P)

    return x
\end{lstlisting}

\vspace{0.18em}
\noindent\hrule height 0.35pt

\end{minipage}
\end{figure}
\subsection{Sparse Sequence Parallel (SSP)}
Skiparse Attention is implemented by alternating Token-wise Sparse Attention (TSA) and Group-wise Sparse Attention (GSA). In both TSA and GSA, each subsequence has a length of $\frac{1}{k}$ of the original sequence and performs global attention. In the non-parallel setting, the transformation from the original pattern to either sparse pattern, as well as the conversion between the two sparse patterns, can be implemented with a single rearrange operation, as shown in Alg.~\ref{alg:ssp}.

For the parallel implementation of Skiparse Rearrange, we employ Sparse Sequence Parallel (SSP), as specified by the $\text{sparse\_sequence\_parallel}$ function in Alg.~\ref{alg:ssp} and Fig.~\ref{fig:method_ssp}. Because Skiparse Rearrange generates subsequences of identical length, these subsequences can be sharded across different ranks for parallel computation, naturally maintaining load balance. More importantly, the local-equivalence property of Skiparse Rearrange enables switching between the two sparse patterns without an All-Gather operation. Instead, each rank applies a local rearrange and then uses All-to-All synchronization to obtain the portion of data assigned to that rank after the sparse-pattern transition. Compared with the naive All-Gather--Rearrange--Reshard implementation, this approach reduces the global communication volume from $O(N^2S)$ to $O(NS)$, where $N$ is the size of the communication group and $S$ is the amount of data maintained by each rank. Furthermore, element-wise operations such as LayerNorm and FFN are independent of the sequence dimension, so they require no additional adaptation. Under this design, each block requires only one All-to-All operation during pattern switching. In contrast, Ulysses Sequence Parallel requires All-to-All communication for the query, key, value, and attention output; Sparse Sequence Parallel therefore reduces the communication volume by 75\%.

Sparse Sequence Parallel is naturally compatible with Ulysses Sequence Parallel, because it only repartitions the enlarged batch dimension obtained after applying Skiparse Rearrange. When both parallelization strategies are enabled, the hidden states are partitioned according to the boundaries of the smallest repeatable units before they enter the main blocks, thereby preserving the properties of Skiparse Rearrange.

\subsection{Model Structure}
\begin{figure}[t]
    \centering
    \includegraphics[width=\linewidth]{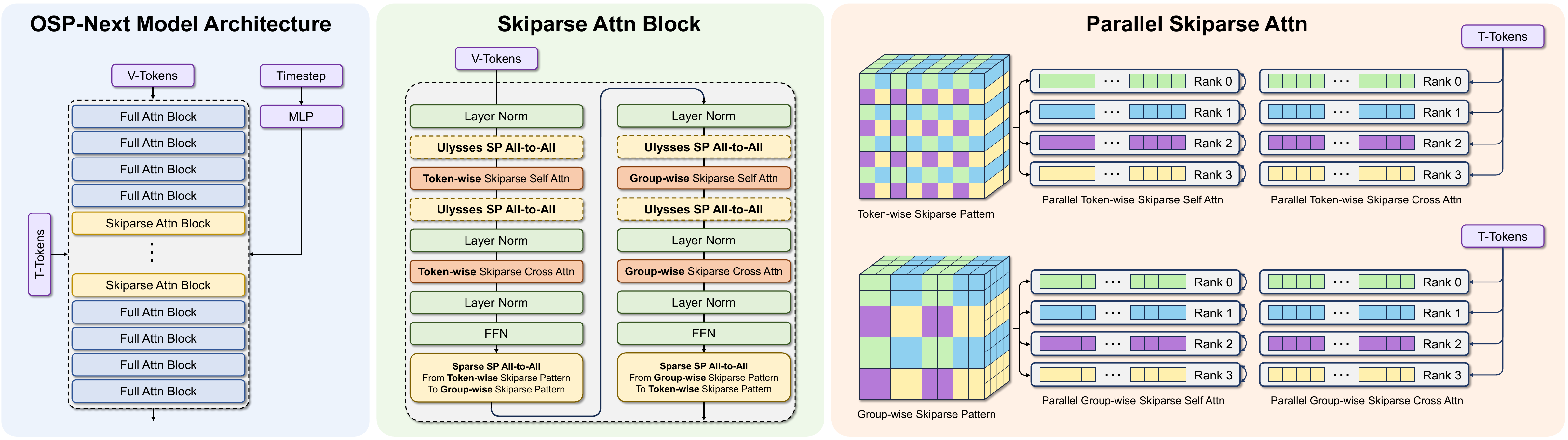}
    \caption{\textbf{Model Architecture.} OSP-Next adopts CrossDiT with a hybrid attention strategy. Specifically, we use full attention blocks in the first $2n$ layers and the last $2n$ layers to better capture fine-grained features, while using Skiparse Attention blocks in the remaining middle layers to maximize runtime efficiency. When computing self attention, the model processes all subsequences in parallel, reducing the computational complexity to $\frac{1}{k}$ of full attention, where $k$ denotes the sparse ratio. When SSP is enabled, each rank holds a subset of the subsequences. When Ulysses SP and SSP are enabled together, each rank obtains the local heads of the corresponding subsequences through an All-to-All communication before attention. Element-wise operations, such as LayerNorm and FFN, are unaffected by the parallelization strategy. During block forward, SSP changes the token pattern through a single All-to-All communication.}
    \label{fig:method_model_architecture}
\end{figure}
In Open-Sora Plan v1.3, we adopt a spindle-shaped architecture, where the first and last $2n$ layers use full attention Layers and the middle layers use Skiparse Attention Layers. In Open-Sora Plan v1.5, we use a U-shaped sparsity schedule, where the sparsity level increases from full attention Layers in the outer layers to large sparse attention layers in the inner layers. To balance performance and efficiency, OSP-Next adopts a spindle-shaped structure similar to that of Open-Sora Plan v1.3, as shown in Fig.~\ref{fig:method_model_architecture}.

\subsection{HiF8 Quantization} 
\label{sec:hif8}
\subsubsection{HiF8 Data Format}
HiF8 is a proprietary 8-bit floating-point format developed by Huawei. Unlike conventional FP8 variants, such as E4M3 and E5M2, which use a fixed exponent bit width, HiF8 introduces a variable-width \emph{Dot} field that dynamically allocates the available bits between the exponent and mantissa. This design mitigates the inherent trade-off between dynamic range and representational precision. Its three core properties are as follows.

\textbf{Wide Dynamic Range.} Conventional FP8-E4M3 fixes the exponent to 4~bits with a bias of 7, resulting in a limited exponent span and frequent overflow or underflow in the deep attention layers of DiT-based generative models. By unifying normal and denormal encodings within a coherent scheme, HiF8 extends the effective exponent range to $[-22,,15]$, covering 38 distinct exponent values and approaching the 40-exponent coverage of FP16, $[-24,,15]$. This dynamic range allows HiF8 to quantize activations and gradients that span several orders of magnitude without fine-grained per-token or per-block rescaling, thereby reducing the engineering complexity and memory overhead of low-precision training pipelines.

\begin{figure}[!t]
    \centering
    \includegraphics[width=0.5\linewidth]{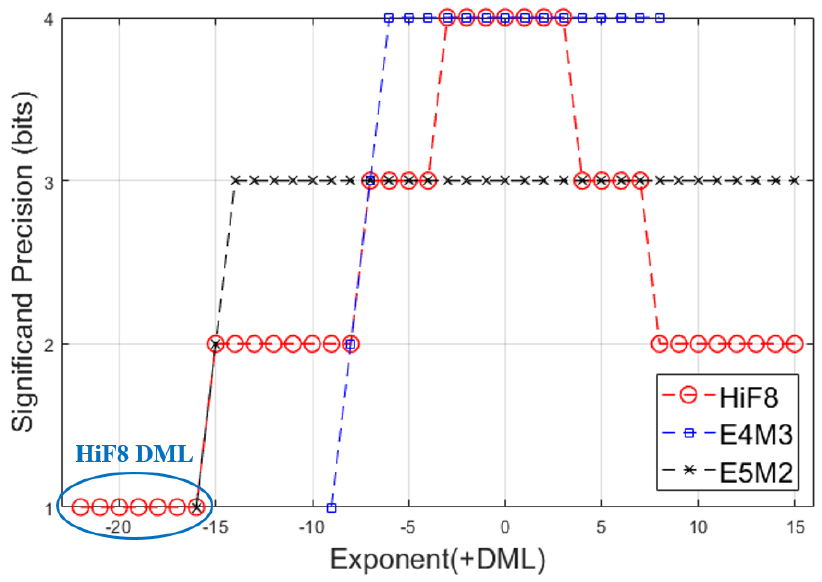}
    \caption{\textbf{Effective mantissa–exponent allocation of 8-bit floating-point formats.}
HiF8 adopts a tapered precision design tailored to the statistical distribution of neural network data. Unlike conventional FP8 formats such as E4M3 and E5M2, whose fixed exponent allocation imposes a trade-off between dynamic range and numerical precision, HiF8 concentrates higher precision in frequently occurring value regions, which typically follow a Gaussian-like distribution in model weights and activations. This enables a better match between numerical representation and deep learning data characteristics.}
    \label{fig:method_hif8}
\end{figure}

\textbf{Tapered Precision.} Conventional FP8 formats face a fundamental trade-off between precision and dynamic range: E4M3 sacrifices dynamic range for higher mantissa precision, whereas E5M2 makes the opposite compromise. Neither configuration fully satisfies both requirements of neural-network training. As shown in Fig.~\ref{fig:method_hif8}, HiF8 addresses this trade-off by progressively reducing the mantissa bit width as the magnitude of a value increases: it preserves a 3-bit mantissa within the central exponent range $[-3,,3]$ and gradually reduces it to a 1-bit mantissa at the extremes of the representable range. This tapered precision profile aligns well with the approximately Gaussian distribution of neural-network weights and activations, concentrating encoding capacity in the high-density region of the numerical distribution while reducing precision only for low-frequency values.

\textbf{Redundancy-Free Encoding.}
The exponent field of HiF8 uses sign-magnitude encoding with an implicit leading 1 bit in the magnitude. This design ensures that the representable exponent ranges of fields with different widths are strictly non-overlapping and that the entire exponent space $[-22,,15]$ is covered without gaps or duplicates. In contrast, conventional FP8 formats inherited from IEEE~754 include redundant code points, such as dual representations of $\pm 0$ and multiple NaN encodings, which waste code points within the already constrained 8-bit budget. In HiF8, every code point maps to a unique numerical value, achieving full encoding utilization and providing a stronger foundation for quantization accuracy under tight bit-width constraints.

\subsubsection{Quantization Scheme}

Mixed-precision training requires a \emph{scaling factor} to map high-precision tensors, such as BF16 and FP32 tensors, into the representable range of HiF8. Let $A_{\max} = \max |\mathbf{X}|$ denote the maximum absolute value of a data block $\mathbf{X}$, and let $\mathrm{HiF8}_{\max}$ denote the largest representable value in the HiF8 format. In this work, $\mathrm{HiF8}_{\max}$ is set to 15 for forward computation and 224 for backward computation, respectively, let $\epsilon$ denote a small positive constant for numerical stability. The scaling factor and quantization step are defined as follows:
\begin{equation}
\mathrm{Scale} = \frac{\mathrm{HiF8}_{\max} }{A_{\max} + \epsilon}
\end{equation}

\begin{equation}
\mathbf{X}_{\mathrm{scaled}} = \mathbf{X} \times \mathrm{Scale}.
\end{equation}

\textbf{Current Scaling.}
We adopt a \emph{current scaling} strategy. At each training iteration, the full input tensor $\mathbf{X}$ is traversed to compute the real-time $A_{\max}$, from which $\mathrm{Scale}$ is derived and immediately applied before downstream computation, such as MatMul. This strategy ensures that the scaling factor remains consistent with the actual data distribution of the current step, thereby eliminating the risk of overflow caused by stale statistics.

\textbf{Per-Tensor Quantization Granularity.}
Thanks to the wide dynamic range of HiF8, a single per-tensor scale is sufficient for the entire tensor, without requiring the per-block granularity, such as $128!\times!128$ tiles, that is typically needed for FP8 formats. Per-tensor quantization substantially reduces the storage and memory-bandwidth overhead associated with scale metadata, simplifies the control logic, and improves end-to-end training throughput.

\subsection{Mix-GRPO For the Sparse-Quant Model}
\begin{figure}[t]
    \centering
    \includegraphics[width=\linewidth]{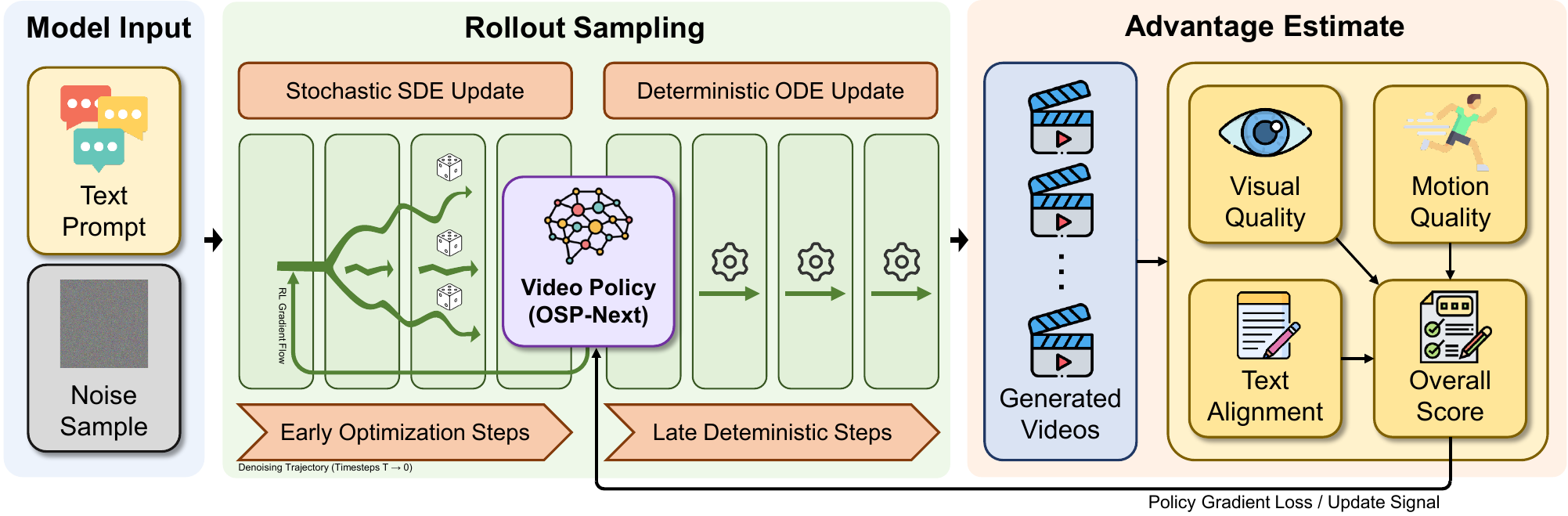}
    \caption{\textbf{Overview of the pipeline for OSP-Next RL post-training.} The policy performs denoising with a hybrid sampling strategy, using stochastic SDE updates in the early steps and deterministic ODE updates in the late steps. The generated videos are then evaluated by VideoAlign along three dimensions—visual quality, motion quality, and text alignment—to produce the training signal for policy optimization.}
    \label{fig:method_osp_rl}
\end{figure}

\subsubsection{Mix-GRPO}

In the post-training stage of the sparse-quant model, we employ a hybrid ODE-SDE denoising strategy following Mix-GRPO~\citep{li2025mixgrpo}, as shown in Fig.~\ref{fig:method_osp_rl}. Although Skiparse-2D Attention and HiF8 quantization substantially improve efficiency, converting a full attention model into a sparse attention model can still introduce a quality gap. We use Mix-GRPO to further align the fine-tuned sparse model with video-generation preferences while preserving the efficiency advantages of OSP-Next.

Conventional flow-based GRPO methods, such as FlowGRPO~\citep{liu2026flow}, formulate the full denoising trajectory as a Markov decision process (MDP). During rollout, SDE sampling is applied at every denoising step to introduce stochastic exploration, followed by policy optimization over the complete state-action sequence. This design is effective for preference alignment but expensive for video generation, because each update requires trajectory-level sampling and policy-ratio computation across all denoising steps. It also mixes gradients from different stages of the generation process: early steps mainly determine global layout and motion structure, whereas late steps focus on local texture and visual details. Jointly optimizing all steps can therefore lead to unfocused updates and unnecessary computation.

Mix-GRPO addresses this issue by combining ODE and SDE sampling. This design leverages an inherent property of diffusion models, namely the ability to introduce more randomness during the early denoising stages. Accordingly, we apply SDE sampling to the initial denoising steps and switch to deterministic ODE sampling in the later stages. In this way, Mix-GRPO shortens the effective MDP horizon, confines stochastic exploration to the optimized timesteps, reduces the number of denoising steps requiring policy-gradient updates, and improves the training efficiency of the overall RL pipeline.

For a deterministic reverse probability flow ODE, the denoising dynamics can be written as:
\begin{equation}
\frac{d\mathbf{x}_t}{dt}
= f(\mathbf{x}_t, t)
- \frac{1}{2} g^2(t) \nabla_{\mathbf{x}_t} \log q_t(\mathbf{x}_t).
\end{equation}

According to the Fokker-Planck equation, this ODE has an equivalent probability flow SDE that maintains the same marginal distribution at each time step:
\begin{equation}
\frac{d\mathbf{x}_t}{dt}
= f(\mathbf{x}_t, t)
- g^2(t) \nabla_{\mathbf{x}_t} \log q_t(\mathbf{x}_t)
+ g(t)\frac{d\mathbf{w}}{dt}.
\end{equation}

Mix-GRPO mixes the two forms by using SDE sampling on the optimized timesteps and ODE sampling otherwise:
\begin{equation}
d\mathbf{x}_t =
\begin{cases}
\left[f(\mathbf{x}_t, t) - g^2(t)\mathbf{s}_t(\mathbf{x}_t)\right]dt + g(t)d\mathbf{w}, & \text{if } t \in S, \\
\left[f(\mathbf{x}_t, t) - \frac{1}{2}g^2(t)\mathbf{s}_t(\mathbf{x}_t)\right]dt, & \text{otherwise}.
\end{cases}
\end{equation}
where we define the score function as
\begin{equation}
\mathbf{s}_t \triangleq \nabla_{\mathbf{x}_t} \log q_t(\mathbf{x}_t).
\end{equation}

Compared with FlowGRPO, this mixed SDE-ODE strategy provides two practical advantages for OSP-Next. First, it reduces optimization overhead because only a subset of timesteps contributes to the GRPO loss. Second, it excludes the ODE portions from the RL objective, thereby reducing unnecessary stochastic sampling and lowering rollout cost. These properties are especially important for video models, where each trajectory contains long spatio-temporal sequences and reward evaluation is substantially more expensive than in image generation.

For OSP-Next, Mix-GRPO therefore serves as an efficient preference-alignment mechanism for the sparse-quant model. It helps recover the performance drop caused by sparse fine-tuning from the pretrained full attention model and by quantization, while keeping the training pipeline compatible with Skiparse-2D Attention, Sparse Sequence Parallelism, and HiF8 simulation. Empirically, this post-training stage provides stable gains for sparse video generation and brings OSP-Next closer to the full attention baseline.

\subsubsection{Reward Model}
We adopt VideoAlign as the reward model. Since VideoAlign~\citep{liu2026improving} is natively trained on videos, it reduces the evaluation gap in video scoring compared with image-trained models such as CLIP~\citep{clip}. In our implementation, VideoAlign provides reward dimensions including visual quality, motion quality, text alignment, and an overall score. We use the overall VideoAlign score as the scalar reward for Mix-GRPO optimization.
\section{Experiment}
\label{sec:experiment}
\subsection{Training Setup}
\textbf{Dataset.} We use an internally collected dataset of high-quality videos and apply the same filtering strategy as described in Open-Sora Plan. We discard videos that are overly static or have excessively low resolution, and train with a fixed duration of 81 frames and a fixed resolution of $720 \times 1280$.

\textbf{Training.} 
The weights of OSP-Next are initialized from the Wan2.1-T2V-14B~\citep{wan2.1} model. We use a configuration with eight full attention blocks, with four placed at the beginning and four at the end, while the middle 32 layers use Skiparse Attention blocks. Skiparse-2D Attention is adopted with a sparse ratio of 2. During training, we use SSP with a parallelism degree of 4. We adopt Fully Sharded Data Parallel (FSDP) as the training strategy and use FSDP-EMA during training, with the EMA decay set to 0.999. We train the model with the AdamW optimizer, using a learning rate of $0.00002$, a weight decay of $0.001$, and a global batch size of 24. The SFT weights are obtained after 6000 training steps. For the Mix-GRPO post-training stage, we adopt an SDE-ODE hybrid sampling strategy with a total of 25 sampling steps, where the first 10 steps use SDE and the remaining 15 steps use ODE. We apply LoRA with a rank of 32 and an alpha of 64 for parameter-efficient fine-tuning. The rollout group size is set to 4. The learning rate is set to $0.0001$, and the KL regularization coefficient $\beta$ is set to $0.004$. OSP-Next-HiF8 is implemented with HiF8 simulation operators and is also initialized from the Wan2.1-T2V-14B model. Its training configuration remains the same as that of OSP-Next.

\textbf{Inference.}
During inference, we use a Discrete Euler sampler with 50 sampling steps, a shift value of 7.0, and classifier-free guidance set to 5.0.

\subsection{Main Results}
\subsubsection{Vbench Scores}
\newcommand{\tabbf}[1]{{\bfseries #1}}

\begin{table}[t]
\caption{\textbf{VBench results.} Compared with the Wan2.1 baseline, OSP-Next achieves a higher total score, demonstrating the effectiveness of the OSP-Next architecture. All metrics reported below are measured in percentages (\%). The abbreviations are defined as follows: SC: Subject Consistency; BC: Background Consistency; TF: Temporal Flickering; MS: Motion Smoothness; DD: Dynamic Degree; AQ: Aesthetic Quality; IQ: Imaging Quality; ObjC: Object Class; MO: Multiple Objects; HA: Human Action; SR: Spatial Relationship; AS: Appearance Style; TS: Temporal Style; OverC: Overall Consistency.}
\label{tab:experiment_vbench}
\centering
\setlength{\tabcolsep}{3.5pt}
\resizebox{\linewidth}{!}{%
\begin{tabular}{lccc|cccccccccccccccc}
\toprule
\textbf{Model} & \textbf{Total} & \textbf{Quality} & \textbf{Semantic} & \textbf{SC} & \textbf{BC} & \textbf{TF} & \textbf{MS} & \textbf{DD} & \textbf{AQ} & \textbf{IQ} & \textbf{ObjC} & \textbf{MO} & \textbf{HA} & \textbf{Color} & \textbf{SR} & \textbf{Scene} & \textbf{AS} & \textbf{TS} & \textbf{OverC} \\
\midrule
Wan2.1-T2V-14B & \underline{83.69} & \tabbf{85.59} & 76.11 & \tabbf{97.52} & \tabbf{98.09} & \tabbf{99.46} & \tabbf{98.30} & \underline{65.46} & 66.07 & 69.43 & 86.28 & 69.58 & 95.40 & \underline{88.59} & \underline{75.39} & 45.75 & \tabbf{22.64} & 23.19 & 25.91 \\
OSP-Next & \tabbf{83.73} & \underline{85.19} & \underline{77.86} & 96.30 & 96.95 & \underline{98.49} & 98.09 & \tabbf{70.83} & \tabbf{66.91} & \underline{69.64} & \underline{90.10} & \underline{77.48} & \tabbf{97.20} & 87.05 & 73.65 & \tabbf{49.39} & \underline{21.93} & \tabbf{24.05} & \tabbf{26.34} \\
OSP-Next-HiF8 & 83.29 & 84.57 & \tabbf{78.21} & \underline{96.61} & \underline{97.10} & 98.40 & \underline{98.26} & 59.44 & \underline{66.44} & \tabbf{70.85} & \tabbf{90.87} & \tabbf{77.70} & \underline{96.80} & \tabbf{90.11} & \tabbf{78.36} & \underline{48.20} & 20.96 & \underline{24.01} & \underline{26.26} \\
\bottomrule
\end{tabular}%
}
\end{table}
We adopt VBench\citep{vbench} as the quantitative metric for assessing model generation quality. As reported in Tab.~\ref{tab:experiment_vbench}, OSP-Next obtains a total VBench score of 83.73\%, surpassing the Wan2.1 baseline and demonstrating the effectiveness of the OSP-Next architecture. Moreover, OSP-Next-HiF8, which uses 8-bit precision, achieves a total score of 83.29\%, keeping the gap from the baseline below 0.4\%. These results suggest that HiF8 precision effectively preserves the capability of the pretrained model while validating the feasibility of joint training between HiF8 and the sparse model.
\subsubsection{Visual Results}
\begin{figure}[!t]
    \centering
    \includegraphics[width=\linewidth]{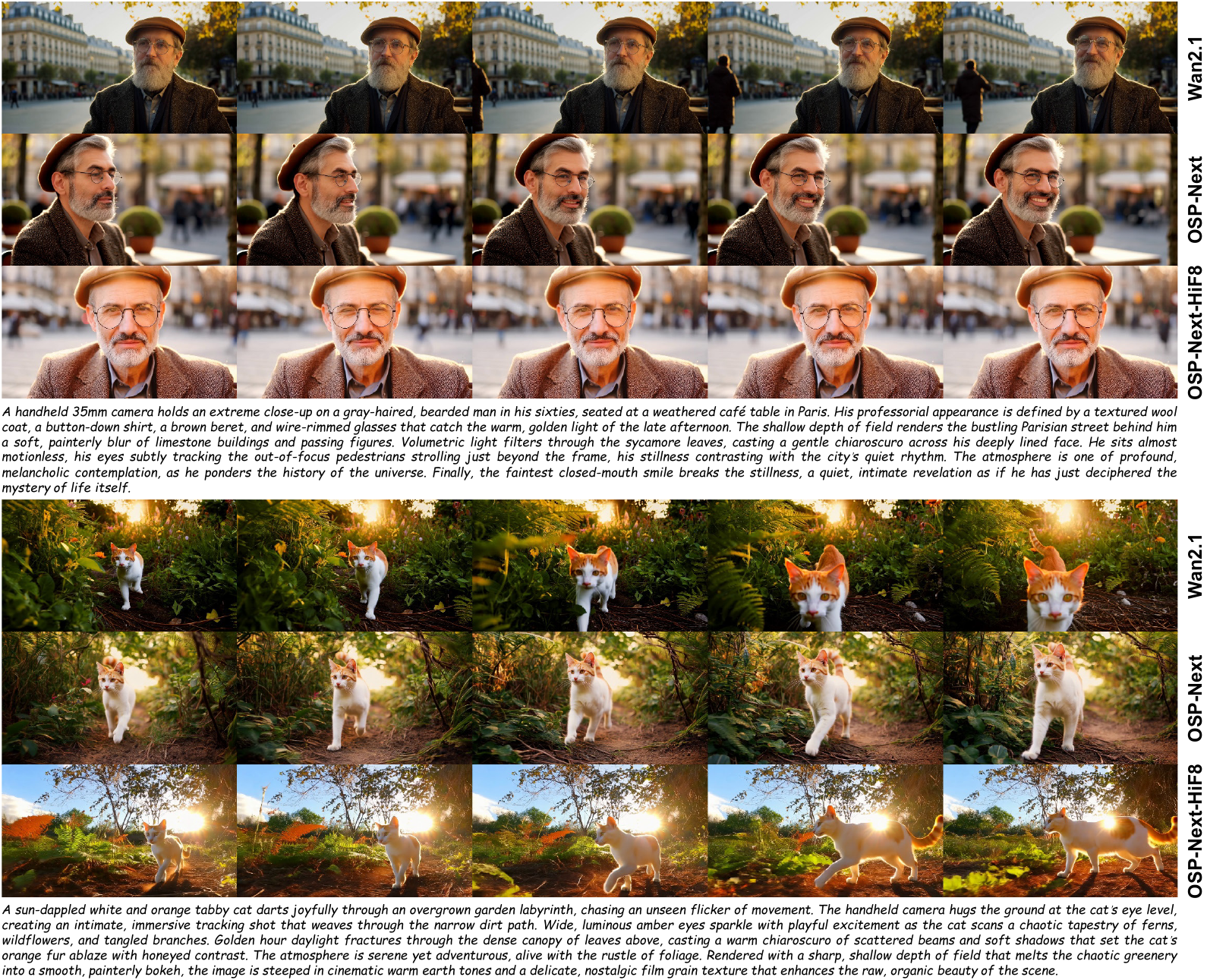}
    \caption{\textbf{Generation quality of the OSP-Next series with the Wan2.1 baseline.} By exploiting video locality through Skiparse-2D Attention and benefiting from the hybrid attention architecture, OSP-Next achieves generation quality that is highly comparable to the Wan2.1 baseline. In addition, due to the dynamic precision of the HiF8 quantization format, OSP-Next-HiF8 shows almost no quality drop compared with OSP-Next.}
    \label{fig:experiment_main_results}
\end{figure}
We compare the visual generation quality of the OSP-Next series with the Wan2.1 baseline, as shown in Fig.~\ref{fig:experiment_main_results}. Although OSP-Next is a sparse Attention model, its visual quality shows no noticeable difference from the baseline. In addition, owing to the dynamic precision of HiF8 quantization, OSP-Next-HiF8 shows no clear degradation in generation quality compared with OSP-Next. These results demonstrate the effectiveness of joint training with sparse-model fine-tuning and 8-bit quantization.

\subsubsection{Inference Speed}
\newcommand{\NA}{\ensuremath{/}}

\begin{table}[!t]
\caption{\textbf{Inference speed.} Since OSP-Next adopts an any resolution strategy, we evaluate two settings with 81 frames: 720P, where padding and attention masks are present, and 768P, where padding and attention masks are absent. OSP-Next achieves more than 1.42$\times$ speedup on different hardware models under both settings, with better performance when padding is absent. The OSP-Next-HiF8 model achieves a 2.27$\times$ speedup on Ascend 950PR, demonstrating the speed advantage of HiF8 quantization. Here, SDPA denotes \texttt{torch.scaled\_dot\_product\_attention}. {\scriptsize \textit{Since Ascend 950PR resources are currently in limited supply, the  results for this hardware are restricted to one-NPU only.}}}
\label{tab:experiment_infer_speed}
\centering
\setlength{\tabcolsep}{2.2pt}
\renewcommand{\arraystretch}{1.05}
\captionsetup[subtable]{font=scriptsize, labelfont={bf,scriptsize}, skip=2pt}

\begin{subtable}[t]{0.49\linewidth}
\centering
\caption{(T, H, W) = (81, 720, 1280). \textbf{With padding.}}
\label{tab:experiment_infer_speed_left}
\scriptsize
\begin{tabularx}{\linewidth}{l *{4}{>{\centering\arraybackslash}X}}
\toprule
\textbf{\scriptsize Model} & \textbf{\scriptsize \mbox{1 GPU/NPU}} & \textbf{\scriptsize \mbox{2 GPU/NPU}} & \textbf{\scriptsize \mbox{4 GPU/NPU}} & \textbf{\scriptsize \mbox{8 GPU/NPU}} \\
\midrule
\multicolumn{5}{c}{\emph{Nvidia H200}, \emph{\texttt{torch.compile} enabled}, \emph{FlashAttention3}} \\
\midrule
Wan2.1 Baseline & 946s & 501s & 254s & 129s \\
OSP-Next & 620s{\tiny(1.53$\times$)} & 335s{\tiny(1.50$\times$)} & 171s{\tiny(1.49$\times$)} & 91s{\tiny(1.42$\times$)} \\
\midrule
\multicolumn{5}{c}{\emph{Ascend 910C}, \emph{\texttt{torch.compile} disabled}, \emph{SDPA}} \\
\midrule
Wan2.1 Baseline & 2353s & 1197s & 523s & \NA \\
OSP-Next & 1362s{\tiny(1.73$\times$)} & 683s{\tiny(1.75$\times$)} & 367s{\tiny(1.43$\times$)} & \NA \\
\midrule
\multicolumn{5}{c}{\emph{Ascend 950PR}, \emph{\texttt{torch.compile} disabled}, \emph{SDPA}} \\
\midrule
Wan2.1 Baseline & 1219s & \NA & \NA & \NA \\
OSP-Next & 957s{\tiny(1.27$\times$)} & \NA & \NA & \NA \\
OSP-Next-HiF8 & 723s{\tiny(1.69$\times$)} & \NA & \NA & \NA \\
\bottomrule
\end{tabularx}
\end{subtable}
\hfill
\begin{subtable}[t]{0.49\linewidth}
\centering
\caption{(T, H, W) = (81, 768, 1280). \textbf{Without padding.}}
\label{tab:experiment_infer_speed_right}
\scriptsize
\begin{tabularx}{\linewidth}{l *{4}{>{\centering\arraybackslash}X}}
\toprule
\textbf{\scriptsize Model} & \textbf{\scriptsize \mbox{1 GPU/NPU}} & \textbf{\scriptsize \mbox{2 GPU/NPU}} & \textbf{\scriptsize \mbox{4 GPU/NPU}} & \textbf{\scriptsize \mbox{8 GPU/NPU}} \\
\midrule
\multicolumn{5}{c}{\emph{Nvidia H200}, \emph{\texttt{torch.compile} enabled}, \emph{FlashAttention3}} \\
\midrule
Wan2.1 Baseline & 1061s & 561s & 283s & 140s \\
OSP-Next & 645s{\tiny(1.64$\times$)} & 342s{\tiny(1.64$\times$)} & 173s{\tiny(1.64$\times$)} & 92s{\tiny(1.52$\times$)} \\
\midrule
\multicolumn{5}{c}{\emph{Ascend 910C}, \emph{\texttt{torch.compile} disabled}, \emph{SDPA}} \\
\midrule
Wan2.1 Baseline & 2372s & 1168s & 587s & \NA \\
OSP-Next & 1353s{\tiny(1.75$\times$)} & 673s{\tiny(1.74$\times$)} & 354s{\tiny(1.66$\times$)} & \NA \\
\midrule
\multicolumn{5}{c}{\emph{Ascend 950PR}, \emph{\texttt{torch.compile} disabled}, \emph{SDPA}} \\
\midrule
Wan2.1 Baseline & 1338s & \NA & \NA & \NA \\
OSP-Next & 762s{\tiny(1.76$\times$)} & \NA & \NA & \NA \\
OSP-Next-HiF8 & 590s{\tiny(2.27$\times$)} & \NA & \NA & \NA \\
\bottomrule
\end{tabularx}
\end{subtable}

\end{table}
To evaluate the acceleration effect of OSP-Next on different hardware platforms, we compare its inference speed on NVIDIA H200, Ascend 910C, and Ascend 950PR, as shown in Tab.~\ref{tab:experiment_infer_speed}. 

On a single GPU/NPU, OSP-Next achieves more than 1.53$\times$ speedup over Wan2.1 under both \emph{with padding} and \emph{without padding} settings. In the multi-GPU/NPU evaluation, the Wan2.1 baseline consistently uses Ulysses SP. For OSP-Next, the Skiparse Attention blocks use only SSP under the 2-GPU/NPU and 4-GPU/NPU settings, and use both SSP and SP under the 8-GPU/NPU setting, while the full attention blocks consistently use Ulysses SP. OSP-Next also achieves stable acceleration in the multi-GPU/NPU evaluation. Since the \emph{without padding} setting requires no attention mask, it yields larger speedups. In addition, OSP-Next-HiF8 achieves up to 2.27$\times$ speedup on Ascend 950PR, demonstrating the efficiency advantage of combining HiF8 precision with Skiparse-2D Attention.
\subsection{Ablation Study}
\subsubsection{Skiparse-2D Initialization}
\begin{figure}[!t]
    \centering
    \includegraphics[width=\linewidth]{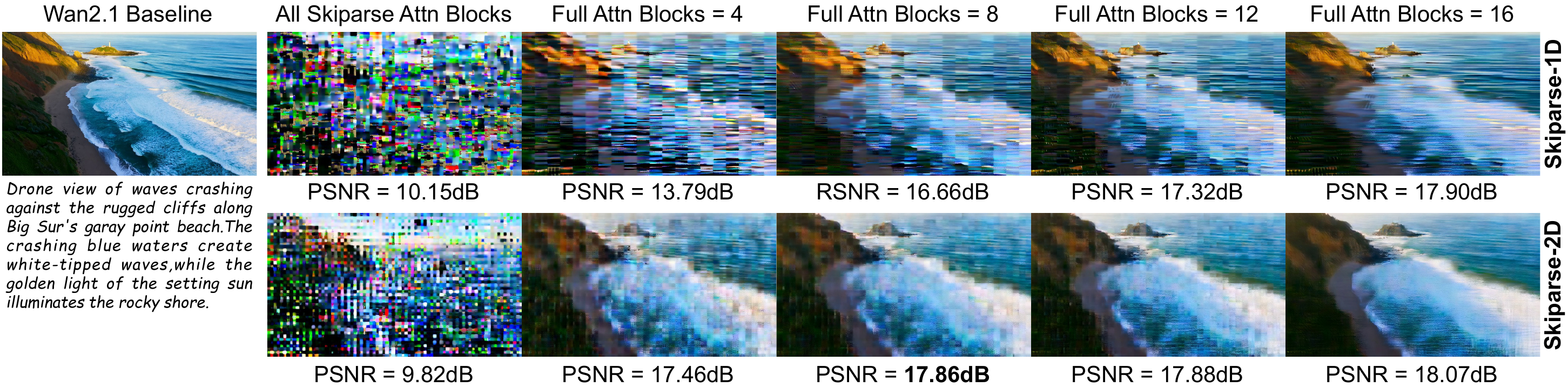}
    \caption{\textbf{Initialization Gap}. OSP-Next adopts Wan2.1-T2V-14B as the initialization weights, whereas an initialization gap remains between the pretrained full attention model and the sparse attention model. To estimate the magnitude of this gap, we compare the zero-shot inference results of the initialized sparse attention model against the baseline. With an increasing number of full attention blocks, the zero-shot inference results of the sparse attention model progressively approach the baseline. Across all settings, Skiparse-2D Attention achieves generation quality that is consistently closer to the baseline than Skiparse-1D Attention. Finally, to balance generation efficiency and generation quality, we choose 8 full attention blocks as the configuration of OSP-Next.}
    \label{fig:experiment_ablation_init}
\end{figure}
OSP-Next is initialized from Wan2.1, which adopts full attention, and therefore faces a transfer gap when adapted to a sparse Attention architecture. For the spindle-shaped architecture used in OSP-Next, we vary the number of full attention blocks and evaluate the corresponding zero-shot generation quality, thereby quantifying the transfer gap between full attention and sparse Attention.

As shown in Fig.~\ref{fig:experiment_ablation_init}, when all blocks are replaced with Skiparse Attention blocks, the zero-shot inference results of the model fail to generate valid videos. This indicates that sparse Attention with predefined rules, such as Skiparse-2D Attention, has difficulty fully inheriting the capability of a pretrained full attention model in the zero-shot setting. OSP-Next alleviates this issue by adopting a spindle-shaped architecture with full attention blocks on both sides. As the number of full attention blocks increases, the zero-shot inference results of the sparse model gradually approach those of the baseline. In addition, the generation quality of the model based on Skiparse-1D Attention lags behind that of the model based on Skiparse-2D Attention, demonstrating the advantage of Skiparse-2D Attention in handling image and video modalities. To achieve a balance between performance and efficiency, OSP-Next uses 8 full attention blocks in its training configuration, with 4 placed at the beginning and 4 placed at the end.

\subsubsection{Reinforce Learning for Sparse Model}
\begin{figure}[!t]
    \centering
    \includegraphics[width=\linewidth]{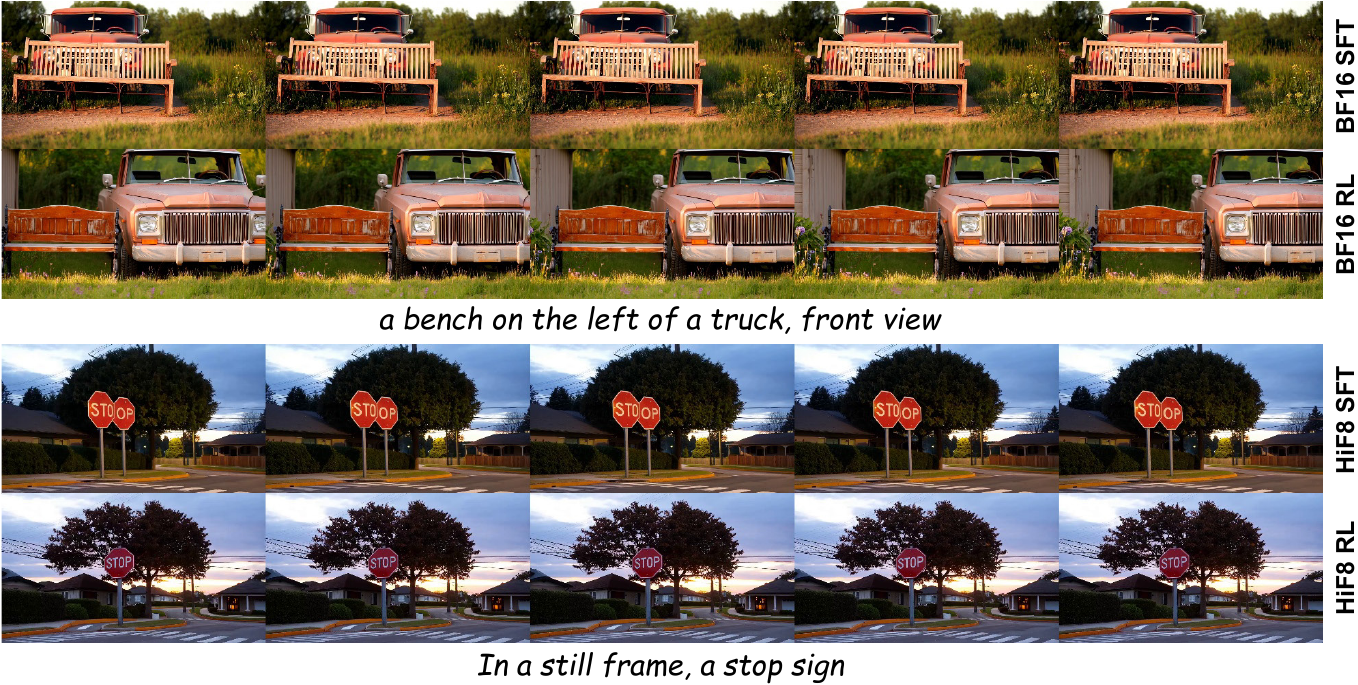}
    \caption{\textbf{Impact of reinforcement learning on the Sparse Attention model.} Adapting a pretrained full attention model to a sparse attention model often results in a degradation of model performance. To mitigate this issue, we use Mix-GRPO for post-training OSP-Next. After post-training, OSP-Next exhibits notable gains in spatial relation understanding, video-text alignment, and related dimensions, indicating that reinforcement learning is effective for the sparse attention model.}
    \label{fig:experiment_ablation_rl}
\end{figure}
Although Skiparse-2D Attention better aligns with the locality of the visual modality and adopts a hybrid attention architecture to better match the capability of the full attention model, the model supervised fine-tuned only on text-video pairs still shows degraded performance compared with the pretrained model. Some prior studies attempt to use the pretrained model as a teacher to align the outputs of the sparse model. However, this approach assumes that the output pattern of the sparse model should be fully consistent with that of the pretrained model, which limits the performance upper bound of the sparse model. Moreover, output-level alignment remains insufficient for capturing the differences between the sparse model and the pretrained model in text-video alignment, aesthetics, and other visual quality-related aspects. 

In OSP-Next, we instead use reinforcement learning to improve the capability of the sparse model. As shown in Fig.~\ref{fig:experiment_ablation_rl}, the model optimized with the Mix-GRPO strategy achieves substantial improvements in spatial relationships, text-video alignment, aesthetics. This demonstrates the effectiveness of reinforcement learning in the post-training stage of the sparse model.

\section{Conclusion}
\label{sec:conclusion}
In this paper, we introduce OSP-Next, a sparse model that employs Skiparse-2D Attention to better leverage the locality inherent in image and video modalities. Motivated by the properties of Skiparse-2D Attention, we further develop Sparse Sequence Parallelism (SSP), a sequence parallelism strategy that is natively aligned with this sparse attention pattern. Compared with Ulysses Sequence Parallelism (SP), SSP reduces communication volume by 75\% and can be used jointly with Ulysses SP. We additionally study the feasibility of joint training with 8-bit quantization for the sparse model, and the effectiveness of reinforcement learning during sparse-model post-training. Experiments demonstrate that OSP-Next maintains comparable VBench scores and generation quality to the pretrained full attention model, while consistently providing over $1.42\times$ speedup across various settings. Moreover, the 8-bit quantized OSP-Next-HiF8 achieves a $2.27\times$ speedup over the pretrained model.

\clearpage

\bibliographystyle{plainnat}
\setlength{\bibhang}{0pt}
\setlength\bibindent{0pt}
\bibliography{main}

\clearpage
\appendix
\begin{figure}
    \centering
    \includegraphics[width=\linewidth]{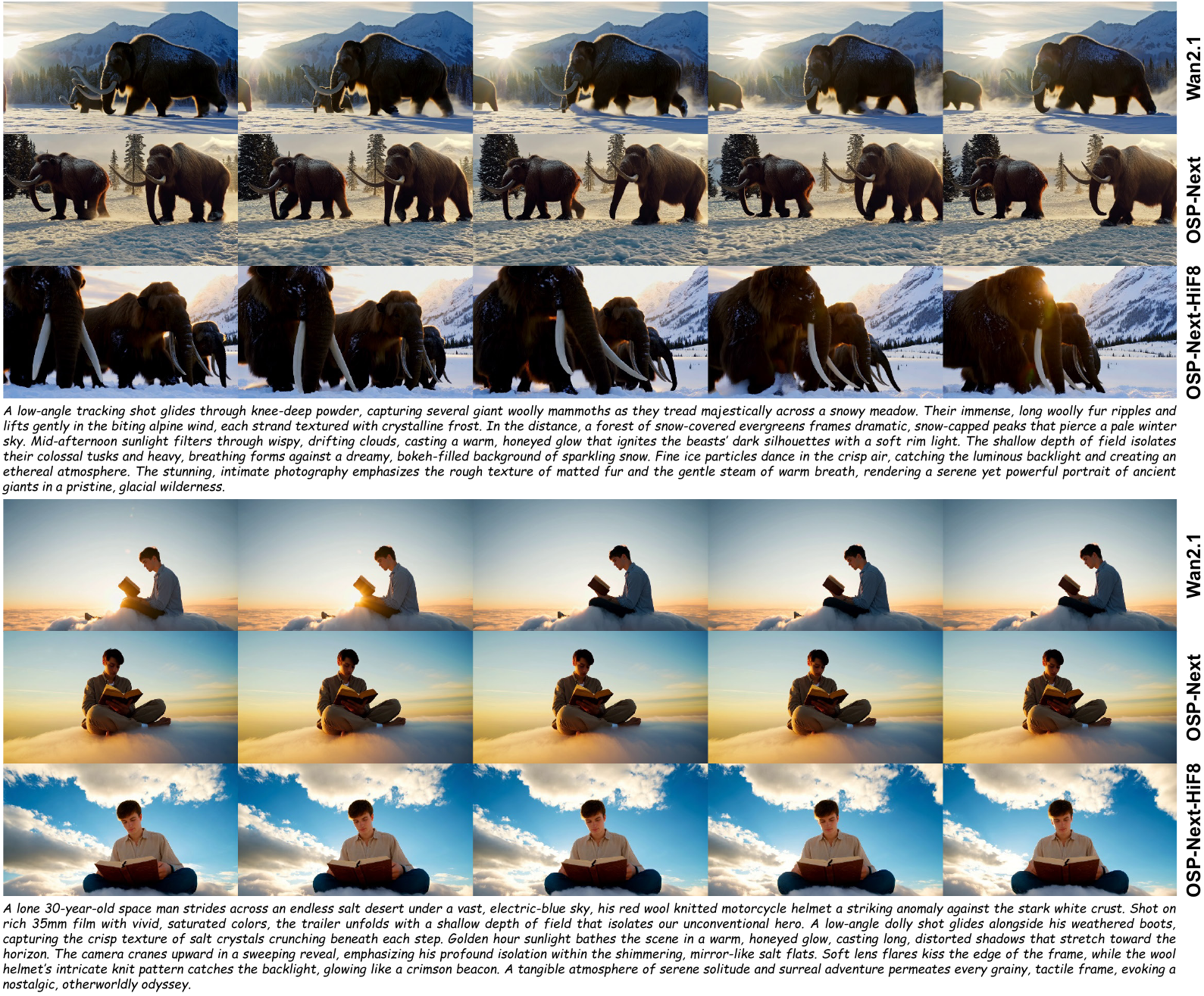}
    \caption{More comparisons between OSP-Next and the Wan2.1 baseline.}
    \label{fig:appendix_1}
\end{figure}
\clearpage
\begin{figure}
    \centering
    \includegraphics[width=\linewidth]{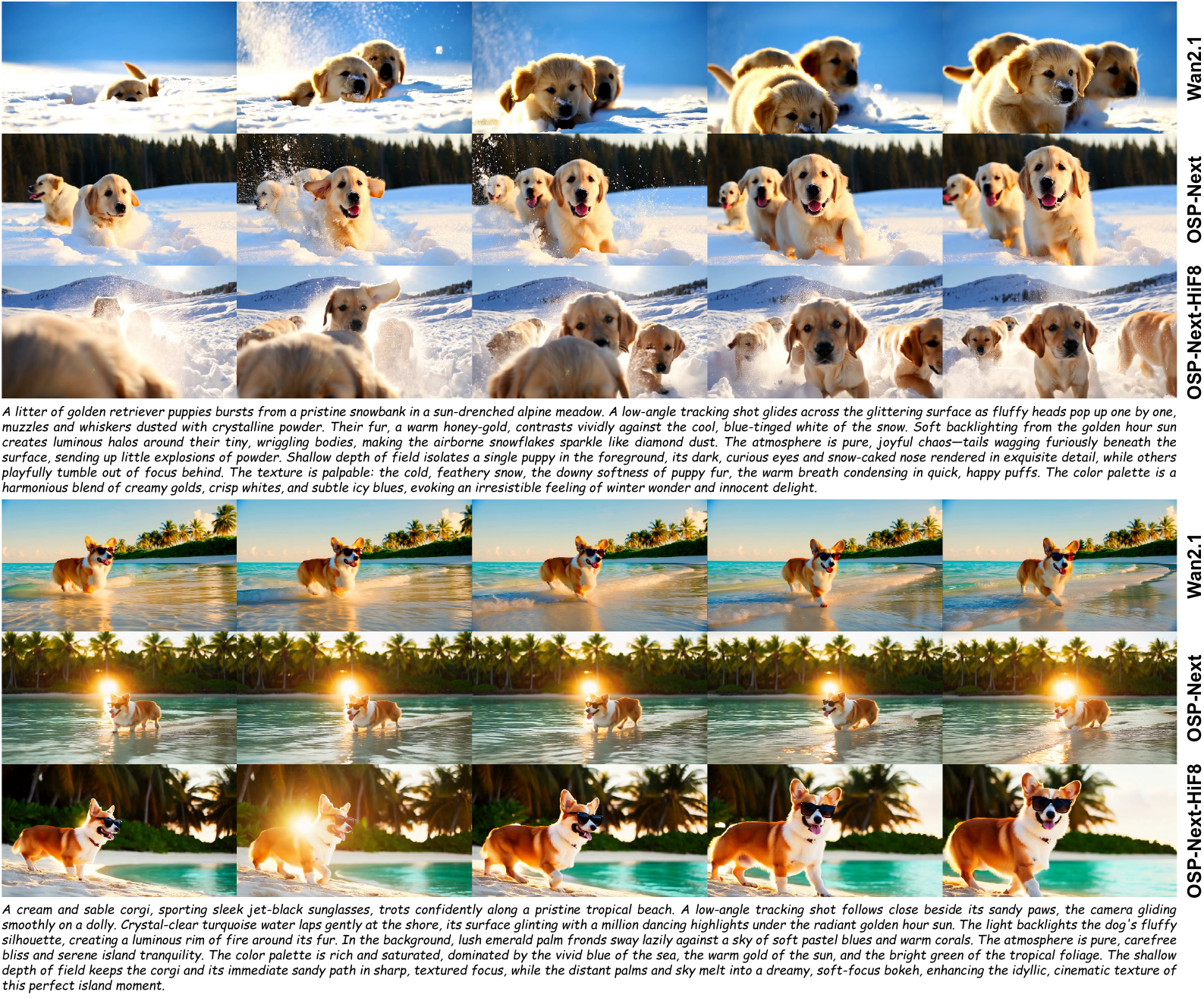}
    \caption{More comparisons between OSP-Next and the Wan2.1 baseline.}
    \label{fig:appendix_2}
\end{figure}
\clearpage
\begin{figure}
    \centering
    \includegraphics[width=\linewidth]{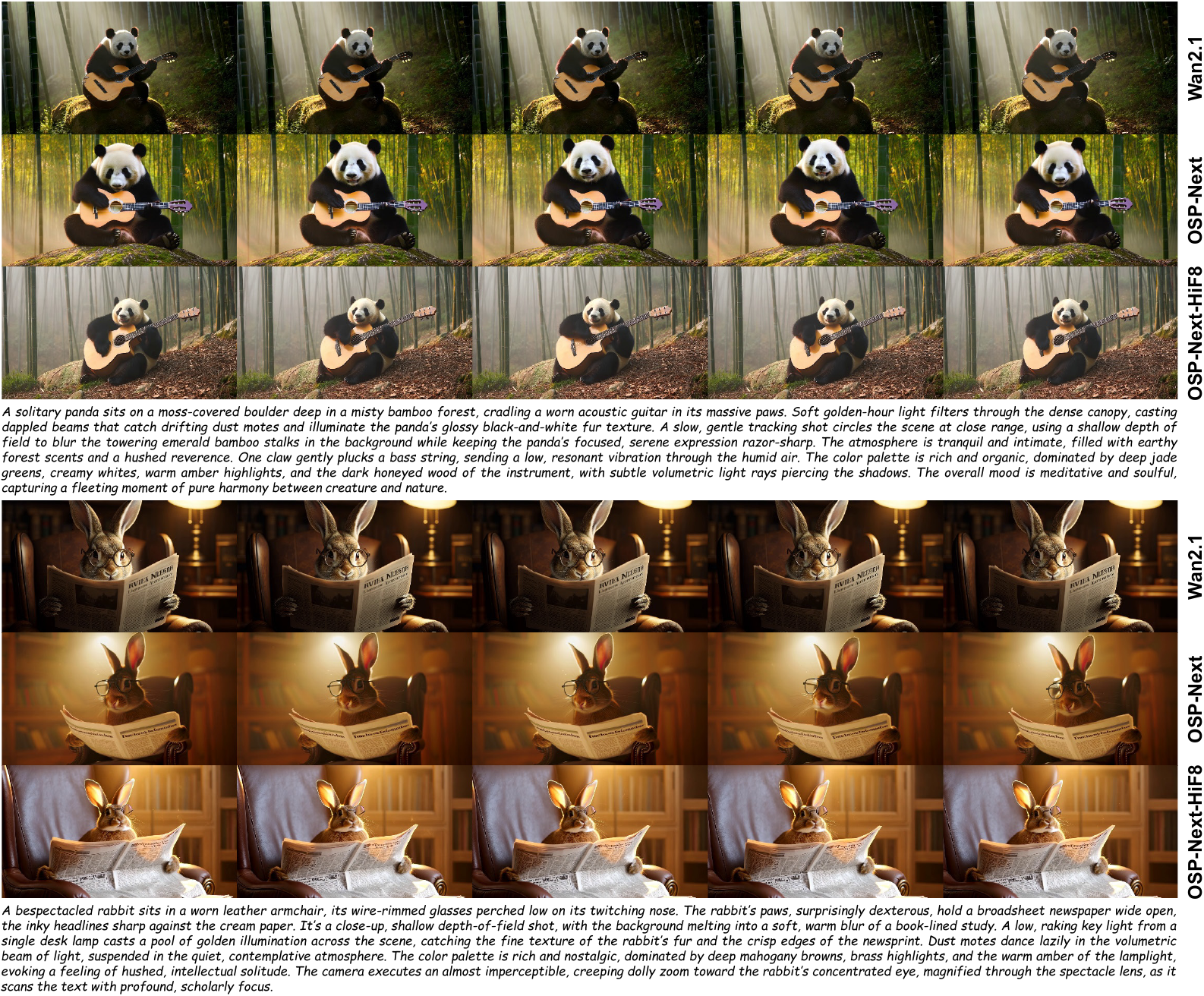}
    \caption{More comparisons between OSP-Next and the Wan2.1 baseline.}
    \label{fig:appendix_3}
\end{figure}
\clearpage
\begin{figure}
    \centering
    \includegraphics[width=\linewidth]{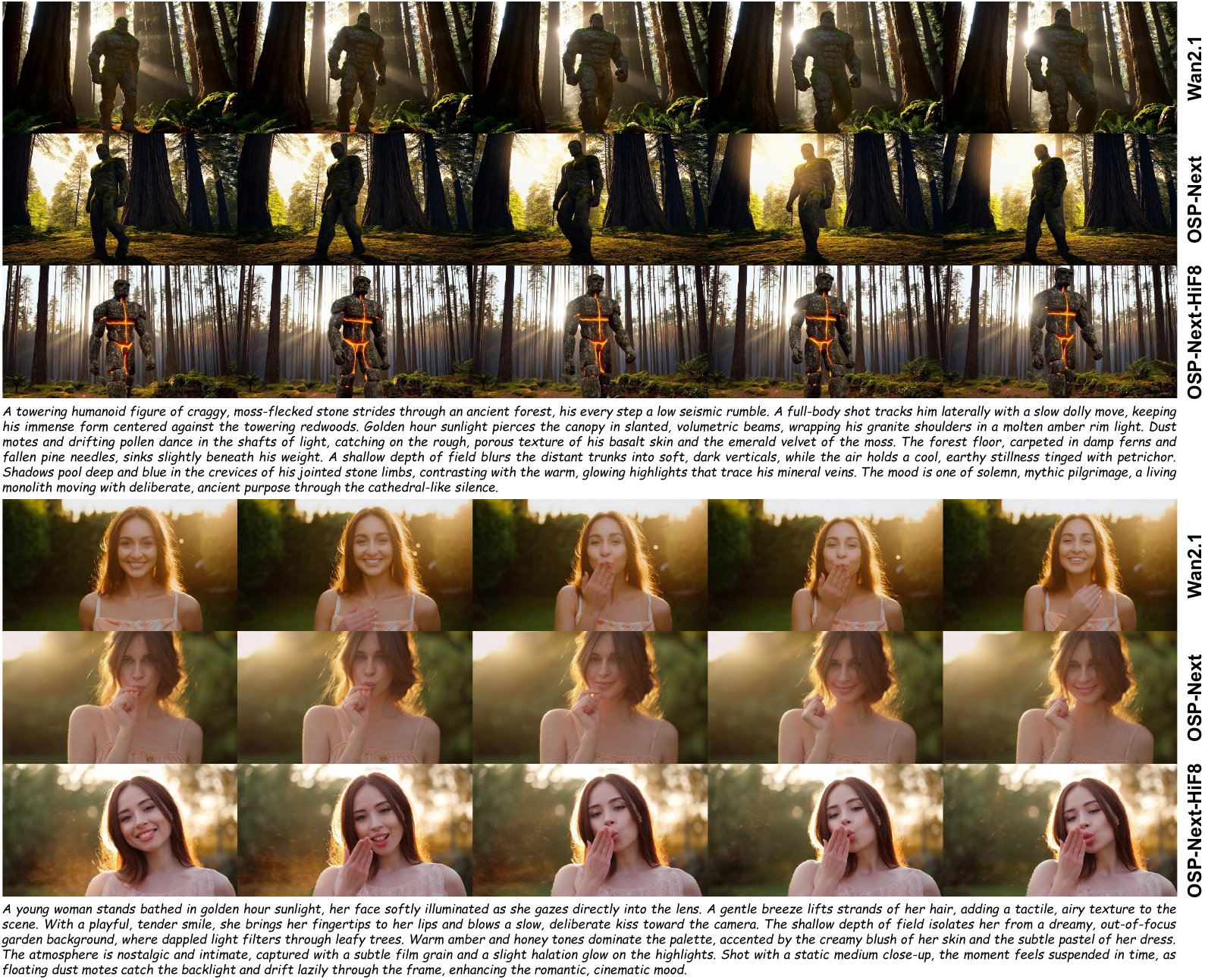}
    \caption{More comparisons between OSP-Next and the Wan2.1 baseline.}
    \label{fig:appendix_4}
\end{figure}
\clearpage
\begin{figure}
    \centering
    \includegraphics[width=\linewidth]{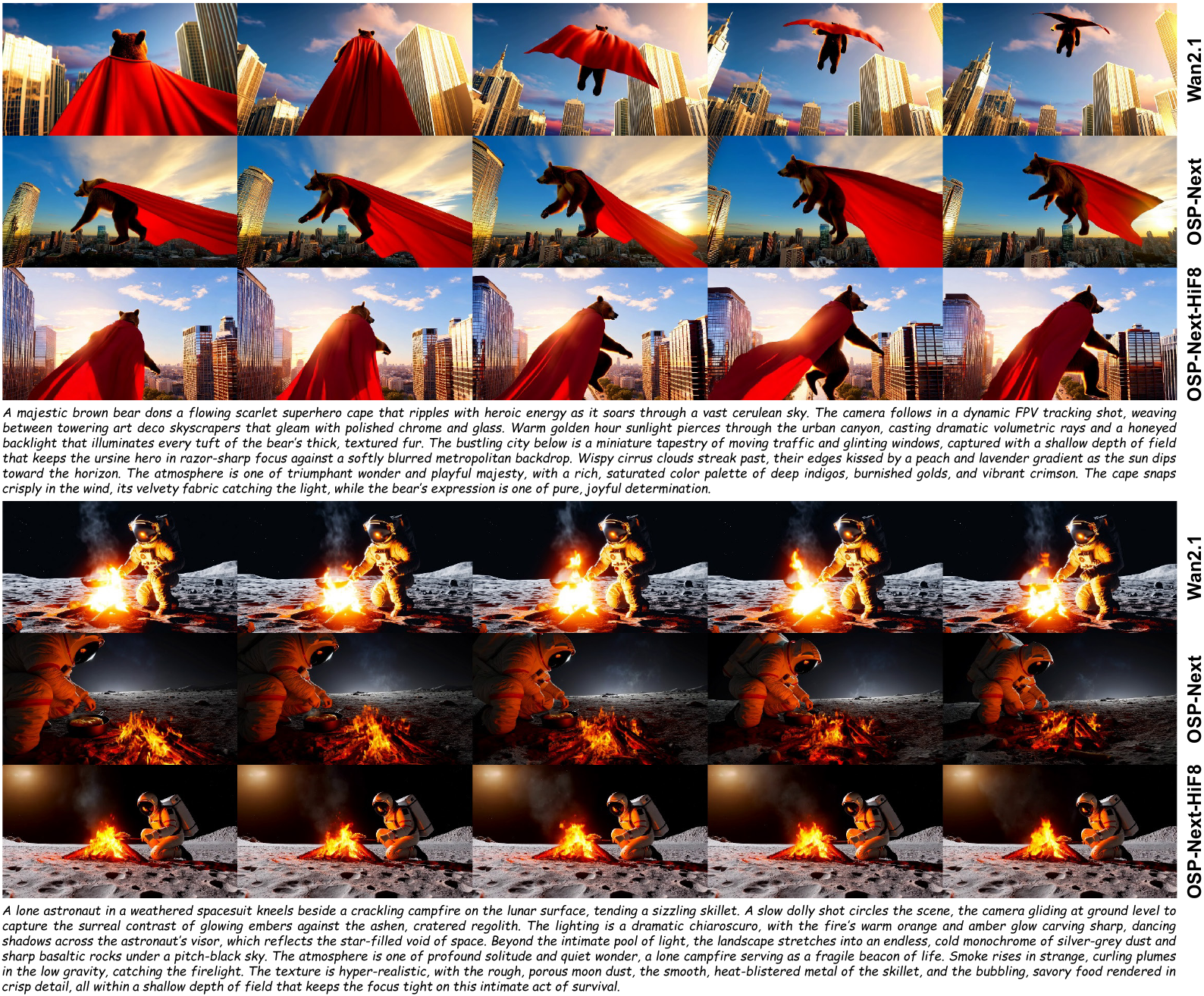}
    \caption{More comparisons between OSP-Next and the Wan2.1 baseline.}
    \label{fig:appendix_5}
\end{figure}
\clearpage
\begin{figure}
    \centering
    \includegraphics[width=\linewidth]{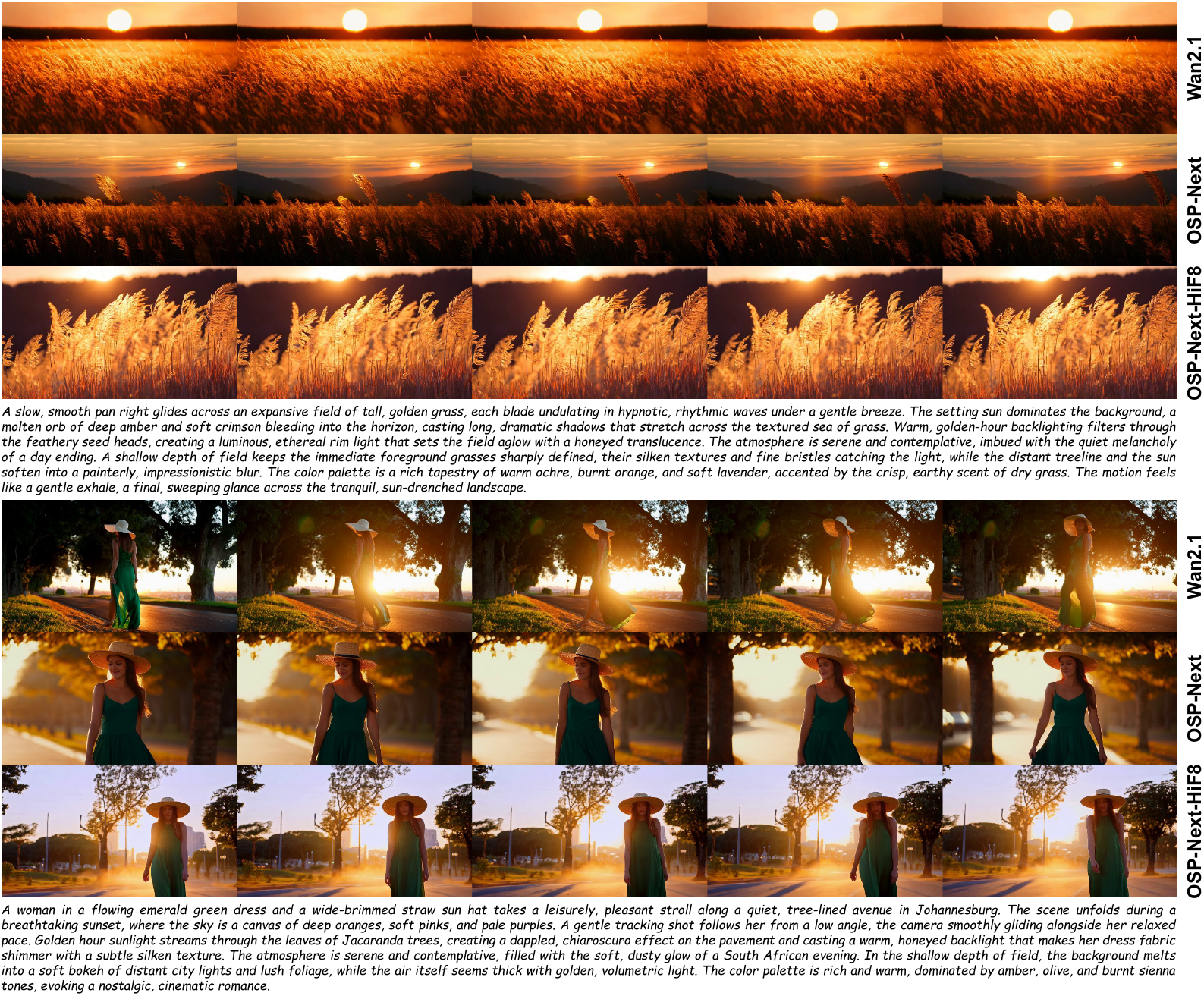}
    \caption{More comparisons between OSP-Next and the Wan2.1 baseline.}
    \label{fig:appendix_6}
\end{figure}

\end{document}